\documentclass[journal]{IEEEtran}

\usepackage{url}  

\usepackage[noadjust]{cite}
\usepackage{enumitem}
\usepackage{amsmath,amssymb,bm}
\usepackage{booktabs}
\usepackage{graphicx,epstopdf,subfigure}
\usepackage{multirow}
\usepackage{tabularx,threeparttable}
\usepackage{array}
\usepackage{xcolor}
\usepackage{comment}
\usepackage[normalem]{ulem} 
\usepackage[colorlinks=true,citecolor=blue,urlcolor=blue]{hyperref}
\usepackage{cite}

\usepackage{fancyhdr}

\newcommand\Tstrut{\rule{0pt}{2.6ex}}         
\newcommand\Bstrut{\rule[-0.9ex]{0pt}{0pt}}   
\newcommand{\revise}[1]{\textcolor{black}{#1}}

\begin{document}

\title{Unpaired Multi-modal Segmentation \\ via Knowledge Distillation}

\author{
	Qi~Dou,~
	Quande Liu,~
	Pheng Ann~Heng,~
	Ben Glocker
	\thanks{Manuscript received July 25, 2019; revised November 12, 2019; accepted December 24, 2019. This work was supported in part by the European Research Council (ERC) under the European Union's Horizon 2020 research and innovation programme (grant No 757173, project MIRA, ERC-2017-STG), and in part by the Hong Kong Innovation and Technology Commission under ITSP Scheme ITS/426/17FP and ITS/311/18FP.}
	\thanks{Q. Dou and B. Glocker are with Biomedical Image Analysis Group, Imperial College London, London, UK. email: \{qi.dou, b.glocker@imperial.ac.uk\}.}
	\thanks{Q. Liu and P. A. Heng are with the Department of Computer Science and Engineering, The Chinese University of Hong Kong, Hong Kong, China. email: \{qdliu, pheng@cse.cuhk.edu.hk\}.}}

\markboth{IEEE Transactions on Medical Imgaging}
{Shell \MakeLowercase{\textit{et al.}}: Bare Demo of IEEEtran.cls for Journals}
	
\maketitle


\begin{abstract}
Multi-modal learning is typically performed with network architectures containing modality-specific layers and shared layers, utilizing co-registered images of different modalities.
We propose a novel learning scheme for unpaired cross-modality image segmentation, with a highly compact architecture achieving superior segmentation accuracy.
In our method, we heavily reuse network parameters, by sharing all convolutional kernels across CT and MRI, and only employ modality-specific internal normalization layers which
compute respective statistics.
To effectively train such a highly compact model, we introduce a novel loss term inspired by knowledge distillation, by explicitly constraining the KL-divergence of our derived prediction distributions between modalities. 
We have extensively validated our approach on two multi-class segmentation problems: i) cardiac structure segmentation, and ii) abdominal organ segmentation.
Different network settings, i.e., 2D dilated network and 3D U-net, are utilized to investigate our method's general efficacy.
Experimental results on both tasks demonstrate that our novel multi-modal learning scheme consistently outperforms single-modal training and previous multi-modal approaches.
\end{abstract}


\section{Introduction}

\IEEEPARstart{A}{natomical} structures are imaged with a variety of modalities depending on the clinical indication. For instance, Computed Tomography (CT) and Magnetic Resonance Imaging (MRI) show cardiac structures with complementary information important for the assessment of heart diseases~\cite{nikolaou2011mri,karim2018algorithms}.
Despite differences between CT and MRI, often a similar analysis is required, such as quantitative assessment via segmentation.
Common practice is to develop a segmentation method, e.g., a convolutional neural network (CNN), separately for CT and MRI data.
Such separate training leaves potentially valuable cross-modality information unused.
By leveraging multi-modal learning, we can exploit shared cross-modality information, possibly making better use of limited datasets and improving overall performance on each modality.

Previous works on multi-modal image segmentation mostly use multi-parametric MRI (e.g., T1, T2, FLAIR).
The inputs to a CNN are paired images, i.e., multi-modal data are acquired from the same patient and co-registered across the sequences.
To learn representations for multi-modal segmentation, early fusion and late fusion strategies are typically utilized.
Specifically, early fusion means concatenating multi-modal images as different channels at the input layer of a network.
This strategy has demonstrated effectiveness on segmenting brain tissue~\cite{chen2018voxresnet,zhang2015deep,moeskops2016automatic} and brain lesions~\cite{fidon2017scalable,kamnitsas2017efficient,valverde2017improving} in multiple sequences of MRI.
For late fusion, each modality has modality-specific layers at an early stage of a CNN.
The features extracted from different modalities are fused at a certain middle layer of the CNN.
The intuition is to initially learn independent features from each modality, and then fuse them at a semantic level.
Briefly speaking, late fusion forms a ``Y"-shaped architecture, as shown in Fig.~\ref{fig:overview} (a).
It has been widely applied for analyzing brain imaging~\cite{nie2016fully}, spinal structures~\cite{li20183d}, prostate cancer~\cite{wang2014computer}, and others.
Recently, more complex multi-modal CNNs have been designed, by leveraging dense connections~\cite{dolz2018hyperdense}, inception modules~\cite{dolz2018ivd} or multi-scale feature fusion~\cite{li2019mman}.
These more complicated models still follow the idea of combining modality-specific and shared layers.

We identify two main limitations in the current multi-modal segmentation literature.
Firstly, input images are typically paired, which requires multiple images from the same patient as well as a registration step at pre-processing.
How to leverage unpaired multi-modal images, e.g, data acquired from different cohorts, still remains unclear.
Secondly, multi-modality is often limited to different sequences of MRI. An arguably more challenging situation of multi-modal learning combining CT and MRI is less well explored.
Due to distinct physical principles of the underlying image acquisition, the very different visual appearance may require new ways to exchange cross-modality information compared to multi-sequence MRI.

To tackle above limitations, studying unpaired multi-modal learning from non-registered CT and MRI has gained some recent interest~\cite{valindria2018multi,zhang2018translating,dou2019pnp,huo2018synseg}.
The very different statistical distributions of CT and MRI makes this a challenging problem in terms of learning shared representations.
As the unpaired images have little pixel-to-pixel coherence, cross-modality relationship only exists in a semantic space.
Valindria et al.~\cite{valindria2018multi} is the closest work to this paper, working on CT/MRI multi-organ segmentation by investigating several dual-stream CNNs, demonstrating a benefit of cross-modality learning of CT and MRI.
The state-of-the-art performance is achieved by a ``X"-shaped network, as shown in Fig.~\ref{fig:overview} (b).
A recent work~\cite{van2019learning} also demonstrate that such an ``X"-shaped model is effective for unsupervised multi-modal learning.
The modality-specific encoder and decoder layers are designed to tackle the distribution shift between two modalities, while the shared middle layers fuse multi-modal representations.

Our paper proposes a novel compact model for unpaired CT and MRI multi-modal segmentation, by explicitly addressing distribution shift and distilling cross-modality knowledge.
We use modality-specific internal feature normalization parameters (e.g., batch normalization layers), while sharing all the convolutional kernels.
Importantly, we further propose to distill semantic knowledge from pre-softmax features.
A new loss term is derived by minimizing the KL-divergence of a semantic confusion matrix, to explicitly leverage the shared information across modalities.
We extensively evaluate our method on two CT and MRI multi-class segmentation tasks, including cardiac segmentation with a 2D dilated CNN and abdominal multi-organ segmentation with a 3D U-Net.
Our method consistently outperforms single-model training and state-of-the-art multi-modal learning schemes on both segmentation tasks. 
The contributions of this work are summarized as follows:
\begin{itemize}
	\item We present a novel, flexible, compact multi-modal learning scheme for accurate segmentation of anatomical structures from unpaired CT and MRI.
	
	\item We propose a new mechanism to distill semantic knowledge from high-level CNN representations. Based on this, we further derive an effective loss function to guide multi-modal learning.
	
	\item We conduct extensive validations on two different multi-class segmentation tasks with 2D and 3D CNN architectures, demonstrating general effectiveness of our method.
\end{itemize}

Code for our proposed approach is publicly available at \url{https://github.com/carrenD/ummkd}.

\section{Related Work}

Before presenting the proposed approach, we review the literature that inspired the design of our multi-modal learning scheme.
The two key aspects are: 1) separating internal feature normalizations for each modality, given the very different statistical distributions of CT and MRI;
2) knowledge distillation from pre-softmax activations, in order to leverage information shared across modalities to guide the multi-modal learning.

\subsection{Independent normalization of CT and MRI}

Representation learning between CT and MRI has attracted increasing research interest in recent years.
Zhang et al.~\cite{zhang2018translating} learn image-to-image translation using unpaired CT and MRI cardiac images.
Dou et al.~\cite{dou2019pnp} present unsupervised domain adaptation of CNNs between CT and MRI for the task of cardiac segmentation using adversarial learning.
Huo et al.~\cite{huo2018synseg} learns a CycleGAN based segmentation model from unpaired CT and MRI, only using segmentation labels from one modality.
In terms of supervised multi-modal segmentation, image style transfer techniques may not be necessary because we have precise annotations to fully supervise the learning process.
To the best of our knowledge, Valindria et al.~\cite{valindria2018multi} is the only paper working on supervised unpaired CT and MRI segmentation so far.
They extensively investigate four different types of dual-stream architectures, showing that a ``X"-shaped architecture obtains the best performance.
This indicates that the distribution shift between CT and MRI heavily affects feature-sharing, requiring modality-specific encoders/decoders.
We hypothesis that if the features from different modalities are better normalized, learning cross-modality representations may become easier.
In literature, independently normalizing features from different domains has demonstrated efficacy for image classification~\cite{bilen2017universal} and life-long learning on multi-modal MRI brain segmentation~\cite{karani2018lifelong}.

\subsection{Knowledge distillation}

The concept of knowledge distillation (KD) originates from Hinton et al.~\cite{hinton2015distilling} for model compression, i.e., transferring what has been learned by a large model to a smaller-scale model using soft-label supervision.
A key aspect that enables KD is to leverage soft labels instead of hard one-hot labels. Temperature scaling is an essential component to allow this by obtaining softer probability distributions across classes, in order to amplify the inter-class relationships.
Previous work has adopted knowledge distillation to address various tasks not limited to original model compression~\cite{hinton2015distilling}, but also a wider scope of scenarios such as domain adaptation~\cite{tzeng2015simultaneous}, life-long learning~\cite{hou2018lifelong}, adversarial attacks~\cite{papernot2016distillation} and self-supervised learning~\cite{lee2018self}.
In medical imaging, the potential of the knowledge distillation technique is promising yet relatively under-explored as far as we know.
Wang et al.~\cite{wang2019segmenting} employ KD for efficient neuronal structure segmentation from 3D optical microscope images with a teacher-student network. 
Kats et al.~\cite{kats2019soft} borrow the concept of KD to perform brain lesion segmentation with soft labels by dilating mask boundaries.
Christodoulidis et al.~\cite{christodoulidis2016multisource} utilize KD for multi-source transfer learning on the task of lung pattern analysis.
With promising results in prior work, we expect an increase of interest in KD.
In this paper, we absorb the spirit of knowledge distillation, and explore how to incorporate effective soft probability alignment with temperature scaling within a novel KD loss derived from processing high-level feature-maps for the task of unpaired multi-modal learning.

\begin{figure*}[t]
	\centering
	\includegraphics[width=0.99\textwidth]{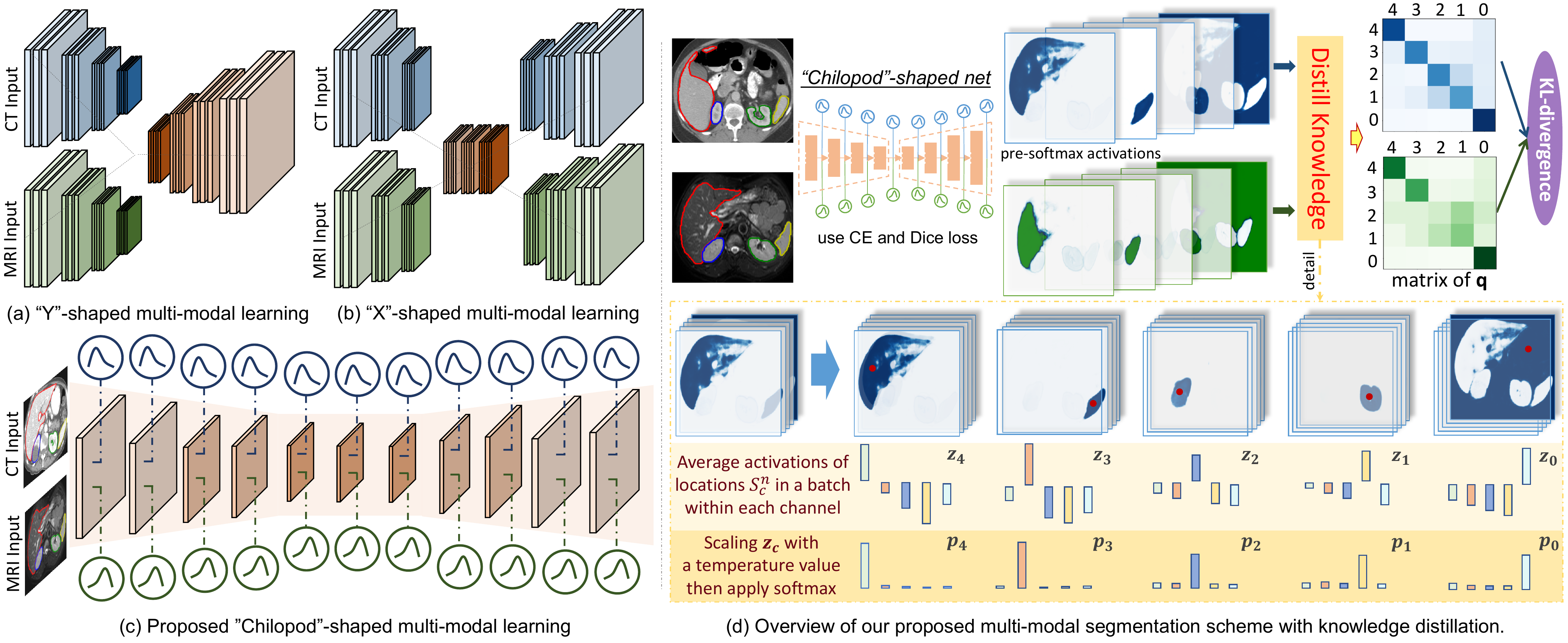}
	\caption{Overview of proposed multi-modal learning scheme for unpaired image segmentation using knowledge distillation. (a) and (b) are the conventional multi-modal learning architectures using modality-specific encoders or decoders, forming a ``X"-shaped or ``Y"-shaped model. (c) Our proposed ``Chilopod"-shaped multi-modal learning architecture, using modality-specific normalization layers, while all convolution kernels are shared. (d) Our proposed method for distilling semantic knowledge from pre-softmax activations, and deriving a pair of confusion matrix for a KL-divergence based loss term.}
	\label{fig:overview}
\end{figure*}

\section{Methods}

An overview of our proposed multi-modal segmentation method is shown in Fig.~\ref{fig:overview} (d). In this section, we first present a compact model design with modality-specific parameters for feature normalization. Next, we show how to distill knowledge from semantic feature-maps and derive the loss term for multi-modal learning. The training procedures are finally described.

\subsection{Separate internal feature normalization}

The central architecture of our proposed learning scheme is a separate normalization for internal activations, to mitigate the discrepancy from different data sources. Our design is very different from previous multi-modal learning methods. Rather than using modality-specific encoders/decoders with early/late fusions, we employ the same set of CNN kernels to extract features for both modalities yielding higher parameter efficiency.
Using these modality-agnostic kernels gives raise to the hope of extracting universal representations that are more expressive and robust.
To achieve this, calibration of the features extracted by the model is important.
Normalizing internal activations into Gaussian distribution is common practice for improving convergence speed and generalization of a network.
Let $x\in S^k_g$ denote the activation in the $k$-th layer, $S^k_g$ is the $g$-th group of activations in the layer for which the mean and variance are computed, the normalization layer is:
\begin{equation}
\hat x = \frac{x-E[\text{x}]}{\sqrt{Var[\text{x}] + \epsilon}}, \quad y = \gamma \hat x + \beta,
\vspace{-1mm}
\end{equation}
where $\gamma$ and $\beta$ denote the trainable scale and shift.
There are different ways to define the activation set of $S^k_g$, e.g, Batch Normalization~\cite{ioffe2015batch}, Instance Normalization~\cite{ulyanov2016instance}, Layer Normalization~\cite{ba2016layer}, and Group Normalization~\cite{wu2018group}.

We employ separate internal normalization for each modality, as the statistics of CT and MRI data are very different and should not be normalized in the same way which otherwise may yield defective features.
For instance, if we have $\mu_{\text{CT}}  = - \mu_{\text{MRI}}$ in a certain layer, then the mean over both would be zero which is meaningless.
Using separate normalization layer for each modality can effectively avoid such problems during multi-modal learning.
Our modality-specific normalization gives raise to a compact multi-modal architecture with minimal extra parametrization $\{\gamma, \beta\}$ forming a ``Chilopod" shape, as illustrated in Fig.~\ref{fig:overview} (c).
More specifically, the normalization layers for different modalities (i.e., CT and MRI) are implemented under separate variable scopes, while the convolution layers are constructed under shared variable scope. In every training iteration, the samples from each modality are loaded separately with sub-groups, and forwarded into shared convolution layers and independent normalization layers to obtain the logits which are employed to calculate the knowledge distillation loss.

\subsection{Knowledge distillation loss}

As we share all CNN kernels across modalities, the encoders are expected to extract universal representations, capturing common patterns such as shape, which may be more robust and discriminative across modalities.
Training, however, becomes more difficult and we find that ordinary objectives, e.g., cross-entropy or Dice loss, are  inadequate.
We propose to explicitly enhance the alignment of distilled knowledge from semantic features of both modalities.

The assumption of KD is that the probabilities from softmax contain richer information than one-hot outputs.
An additional temperature scaling for the pre-softmax activations gives softer probability distributions over classes, which can further amplify such knowledge.
In our approach, we distill semantic knowledge from high-resolution feature maps before softmax, as illustrated in Fig.~\ref{fig:overview} (d).
We average the activations over all locations of each class and compute the soft predictions across all classes.
We describe this process for 2D below, while it can be easily extended to 3D.

We denote the activation tensor before softmax by $M \in \mathbb{R}^{N\times W\times H\times C}$, where $N$ is the batch size, $W$ and $H$ denote width and height, $C$ is number of channels which also equals to the number of classes as $M$ is pre-softmax tensor.
Let $z_{nwhi} \in M$ denote one neuron activation in $M$ with index of $(n,w,h,i)$, and $\mathcal{S}^n_c$ denote the set of $(w,h)$ locations in the $n$-th sample where the pixel's label is class $c$.
Then, for each class $c$, inside each channel of $M$, we distill the knowledge over all the locations which belong to class $c$.
There are $C$ channels in total, one for each class. Hence, we get a $C$-dimensional vector $\textbf{z}_c \! \in \! \mathbb{R}^C$ for class $c$, 
with its $i$-th element $\textbf{z}_c^i$ as averaging $z_{nwhi}$ over all $\mathcal{S}^n_c$ locations in the $i$-th class channel.
Formally, this procedure is represented as:
\begin{align}
\textbf{z}_c^i = \frac{1}{ \sum_{n} |\mathcal{S}^n_c|} \sum_{n}\!\!\!\! \sum_{~~(w,h)\in S^n_c} \!\!\!\! z_{nwhi}.
\end{align}

\begin{figure*}[t]
	\centering
	\includegraphics[width=1.02\textwidth]{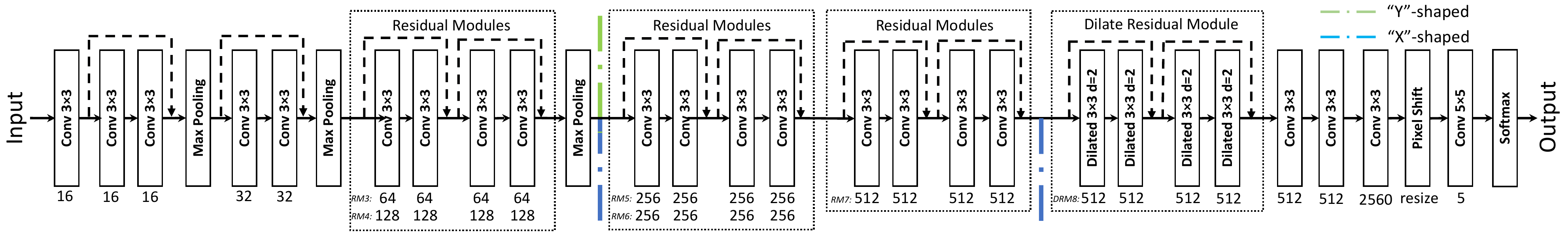}
	\vspace{-6mm}
	\caption{Architecture of the 2D dilated network for multi-class cardiac structure segmentation, following~\cite{DBLP:conf/ijcai/DouOCCH18}. The green and blue lines indicate break/merge points for ``Y"-/``X"-shaped architectures. Best viewed in the zoomed-in view.}
	\label{fig:arch-2d}
\end{figure*}

\begin{figure*}[t]
	\centering
	\includegraphics[width=0.8\textwidth]{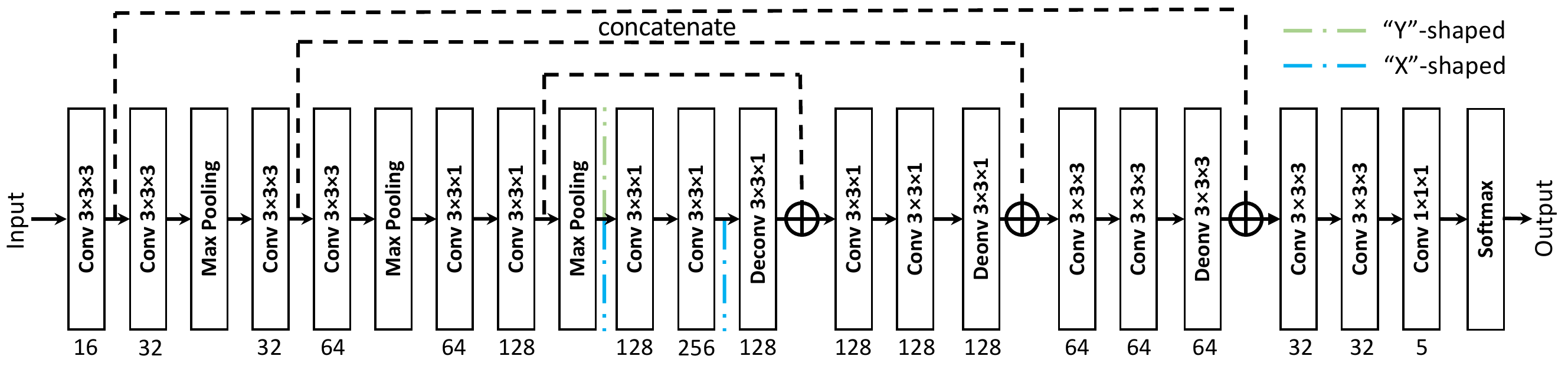}
	\vspace{-3mm}
	\caption{Architecture of the 3D U-Net~\cite{cciccek20163d} for multi-organ segmentation. The black dot lines indicate skip connections. The green and blue lines indicate break/merge points for ``Y"-/``X"-shaped architectures.}
	\label{fig:arch-3d}
\end{figure*}

Next, we compute scaled $\textbf{z}_c$ into a probability distribution $\textbf{p}_c \! \in \! \mathbb{R}^C$ using softmax.
This distilled knowledge $\textbf{p}_c$ aggregates how the network's prediction probabilities for the pixels of class $c$ distribute across all other classes. Temperature scaling is employed as:
\begin{align}
\textbf{p}_c^i = \frac{\exp(\textbf{z}^i_c/T)}{\sum_j \exp(\textbf{z}^j_c/T)},
\end{align}
where $T\!>\!1$ is the temperature scalar~\cite{hinton2015distilling} for softer output to enhance small values. We set $T\!=\!2$ in our experiments, with $T\!=\!1$ being the ordinary softmax.
We empirically observe that the model performance is not very sensitive to this hyper-parameter, within a reasonable range of $T\!<\!10$ considered.

Similarly, we can get an array of distilled semantic knowledge $\textbf{q}=[\textbf{p}_0, \textbf{p}_1, ..., \textbf{p}_{C-1}]$. Each vector element conveys how the model's predictions for pixels of a particular class would distribute across all classes. We can distill such knowledge for each modality, denoted by $\textbf{q}^a$ and $\textbf{q}^b$ for CT and MRI, respectively.
We encourage the network's distilled knowledge from high-level representations of both modalities to be aligned.
Intuitively, if one class in CT is often confused by another, this situation may also happen in MRI.
For instance, as shown in top-right confusion matrix (the blue and green grid planes) of Fig.~\ref{fig:overview} (d), the confusions between class \emph{1} and \emph{2} are more obvious, and this is consistent in CT and MRI under our learning scheme.
These two classes correspond to the left and right kidney (in human-body view) in abdominal images.

In our scheme, both $\textbf{q}^a$ and $\textbf{q}^b$ are updated together during the dynamic training process.
We compute their relative entropy between vectors of each class and design a loss term to minimize their Kullback$-$Leibler (KL) divergence.
Our knowledge distillation based loss term (KD-loss) is as follows:
\begin{equation}
\begin{aligned}
\small
\mathcal{L}_{\text{kd}} =  \frac{1}{C} \sum_c \! \left(\mathcal{D}_{\text{KL}}(\textbf{q}^a_c || \textbf{q}^b_c) + \mathcal{D}_{\text{KL}}(\textbf{q}^b_c || \textbf{q}^a_c) \right), \\ 
\text{where} ~ \mathcal{D}_{\text{KL}}(\textbf{q}^a_c || \textbf{q}^b_c) = \sum \textbf{q}_c^a \log \frac{\textbf{q}_c^a}{\textbf{q}_c^b}.
\end{aligned}
\end{equation}

We compute a symmetric version of KL-divergence between two modalities, and $\mathcal{D}_{\text{KL}}(\textbf{q}^b_c || \textbf{q}^a_c)$ is similar to $\mathcal{D}_{\text{KL}}(\textbf{q}^a_c || \textbf{q}^b_c)$.
By explicitly enhancing alignment of the distilled knowledge, the shared kernels can extract features capturing cross-modality patterns co-existing in CT and MRI at multi-modal learning.

\subsection{Overall loss function and training procedure}

The compact segmentation network is trained with the loss function as follows:
\begin{equation}
\small
\mathcal{L} =  \mathcal{L}_{\text{seg}}^a + \mathcal{L}_{\text{seg}}^b + \frac{\alpha}{2} \mathcal{L}_{\text{kd}} + \eta \left(\| \Theta_{\text{kernel}}\|^2_2 + \| \Theta_{\text{norm}}^a \|^2_2 + \| \Theta_{\text{norm}}^b \|_2^2 \right),
\label{eq:overall_loss}
\end{equation}
where $\mathcal{L}_{\text{seg}}^a$ and $\mathcal{L}_{\text{seg}}^b$ are ordinary segmentation loss for each modality, for which we combine Dice loss~\cite{milletari2016v} and pixel-wise weighted cross-entropy loss in our experiments.
We multiply a scalar $1/2$ for $\mathcal{L}_{\text{kd}}$ to average over the symmetric KL-divergence. 
The $\alpha$ is generally set as 0.5, and we also study this hyper-parameter in ablation experiments.
The last term is a L2 regularizer for shared kernels and modality-specific normalization parameters.
The weight $\eta$ is fixed as $1e\!-\!4$. 

The multi-modal images are sampled to similar voxel-spacing, and normalized to zero mean, unit variance for each modality at the input layer.
In each training iteration, we input half of the batch as CT and the other half as MRI. The images go through the shared convolutional kernels and respective internal normalization layers. 
This can be easily implemented in TensorFlow, by defining the scope of involved variables.
The KD-loss is computed with activation tensors of the samples from each modality. All trainable parameters $\{  \Theta_{\text{kernel}}, \Theta_{\text{norm}}^a, \Theta_{\text{norm}}^b  \}$ are updated together with the loss function in Eq.~\ref{eq:overall_loss} using an Adam optimizer. 
Note that our multi-modal learning scheme is architecture-independent. It can be easily integrated into various existing 2D and 3D CNN models.

\section{Experiments}

We extensively evaluate our multi-modal learning approach on two different multi-class tasks: 1) cardiac structure segmentation; and 2) multi-organ segmentation.
We implement 2D and 3D models with different network architectures to demonstrate the flexibility and general efficacy of our method.

\subsection{Datasets and networks}

\textbf{Task-1}: We perform multi-class cardiac structure segmentation using the MICCAI 2017 Multi-Modality Whole Heart Segmentation Challenge~\cite{zhuang2019evaluation} dataset, which consists of unpaired 20 CT and 20 MRI images from different patients and sites.
The multi-class segmentation includes four structures: left ventricle myocardium (LVM), left atrium blood cavity (LAC), left ventricle blood cavity (LVC) and ascending aorta (AA). 
We crop the heart regions in the images. Both modalities are resampled to a voxel-spacing at around 1.0$\times$1.0$\times$1.0 $mm^3$ with size of 256$\times$256 in the coronal plane. Before inputting to the network, we conduct intensity normalization to zero mean and unit variance separately for each modality.
Each modality is randomly divided into 70\% for training, 10\% for validation and 20\% for testing.

\textbf{Network-1}: For the cardiac segmentation, we implement a 2D CNN with dilated convolutions, following the architecture \revise{used in~\cite{dou2019pnp} which} employ the same dataset.
The inputs are three adjacent slices with the middle slice providing a ground truth mask.
The batch size is set as 8 for each modality. The learning rate is initialized as $1e-4$ and decayed by $5\%$ in every 1000 iterations.
The detailed network architecture is presented in Fig.~\ref{fig:arch-2d}, with size of convolution kernels indicated within the each box. 
The number of channels of feature maps are indicated below each box. 
The normalization layer follows each convolution operation before applying ReLU non-linearity, except for the last knowledge distillation layer before applying softmax.

The layers at which we separate or merge the data streams in the \textit{``Y"-shaped} and \textit{``X"-shaped} architectures are indicated using the colored lines in the architectures. For the \textit{``Y"-shaped} model, layers before green line are separate, and layers after green line are shared between modalities. For the \textit{``X"-shaped} model, layers before the first blue line and after the second blue line are separate, and layers in-between the blue lines are shared between modalities.

\textbf{Task-2}: We perform multi-organ segmentation in abdominal images for liver, spleen, right kidney (R-kdy) and left kidney (L-kdy).
We utilize public CT dataset of~\cite{landman2015multi} with
30 patients (but one case was excluded due to low image quality), 
and our MRI data come from the ISBI 2019 CHAOS Challenge, with 9 cases available at the time we downloaded data.
This enables us to observe how our method performs in the situation when one modality has much fewer samples than the other.
We crop the original CT and MRI images at the areas of multi-organs by excluding the black margins.
Since these two datasets have large variance in voxel-spacing. We resample them into around $1.5 \times 1.5 \times 8.0 ~ \text{mm}^3$, with size of $256\times 256$ in transverse plane.
The images are normalized to zero mean and unit variance for intensities within each modality before inputting to the network.
Again, each modality is randomly divided into 70\% for training, 10\% for validation and 20\% for testing.

\textbf{Network-2}:
We employ a 3D U-Net~\cite{cciccek20163d} wise architecture as shown in Fig.~\ref{fig:arch-3d}, where dot-lines indicate skip connections in the network. The volumetric input has a size of $256 \times 256 \times 8$, considering to contain all organs inside the transverse view and also the constrain from GPU memory. The batch size is set as 4 at training.
Noting that at every training iteration, the balance of the number of samples in a mini-batch for each modality still holds, considering training stability.
The learning rate is initialized as $1e-4$ and decayed by $5\%$ in every 500 iterations.
The indication for \textit{``Y"-shaped} and \textit{``X"-shaped} architectures with green and blue lines are the same as described above for Network-1.

\begin{figure*}[t]
	\centering
	\includegraphics[width=0.98\textwidth]{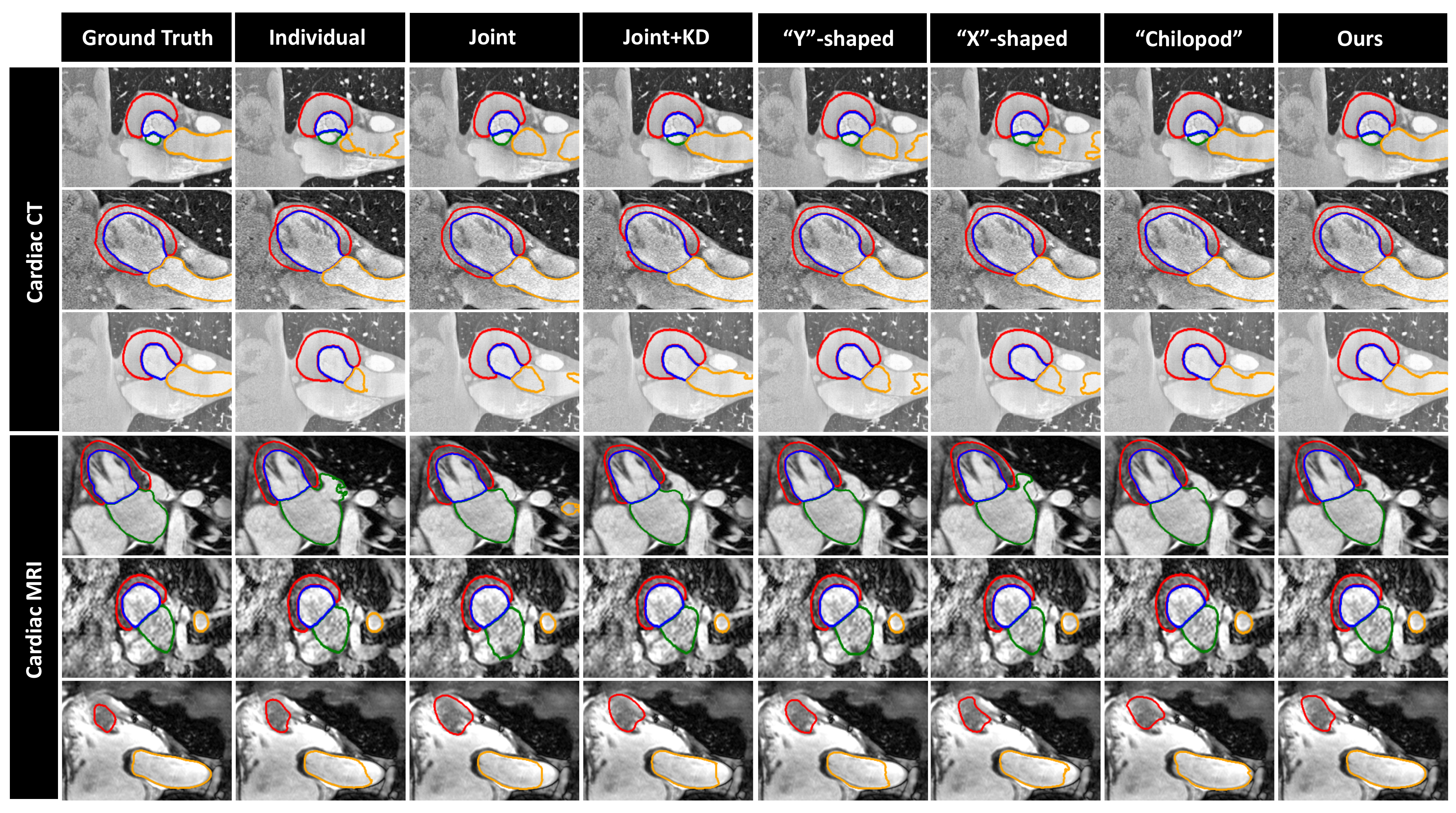}
	\vspace{-3mm}
	\caption{Qualitative comparison of the segmentation results from the seven different settings on cardiac segmentation. The red, yellow, blue and green colors denote the boundary of LVM, AA, LVC and LAC, respectively.}
	\label{fig:fig-cardiac}
\end{figure*}

\begin{table*}[t]
	\centering
	\caption{\small{Multi-modal learning results of cardiac segmentation with 2D network under different settings.}}
	\vspace{-2mm}
	\scalebox{0.83}{
		\begin{tabular}{c|c|ccccc|ccccc|c}
			\toprule
			\multirow{2}{*}{Methods}        &  Param  & \multicolumn{5}{c|}{Cardiac CT} & \multicolumn{5}{c|}{Cardiac MRI} & Overall \Tstrut \Bstrut \\ 
			\cline{3-12}
			& Scale & ~LVM  & ~LAC~ & ~LVC~  &  ~AA~  & Mean   & ~LVM & ~LAC~ & ~LVC~  &  ~AA~ & Mean & ~Mean \Tstrut \\
			\hline
			& & \multicolumn{10}{c|}{\textbf{Dice Coefficient (avg.$\pm$std., $\%$)}} &  \Tstrut \Bstrut \\ 
			\hline
			Payer et al.~\cite{payer2017multi}    &  -      & 87.2$\pm$3.9  & \textbf{92.4$\pm$3.6}  & 92.4$\pm$3.3   & 91.1$\pm$18.4  & 90.8   & ~75.2$\pm$12.1  & ~81.1$\pm$13.8  & 87.7$\pm$7.7   & ~76.6$\pm$13.8  & 80.2  & ~85.5 \Tstrut \\
			Individual                            & 78.64M  & 86.5$\pm$2.0  & 91.7$\pm$2.9           & 90.9$\pm$2.0   & 84.4$\pm$13.0  & 88.4   & 78.7$\pm$4.2    & 84.4$\pm$6.5    & 92.0$\pm$4.0   & \textbf{84.1$\pm$5.4}   & 84.8  & 86.6 \\
			\hline
			Joint                                 & 39.32M  & 84.5$\pm$2.1  & 87.6$\pm$3.5           & 90.2$\pm$2.5   & 92.2$\pm$4.7   & 88.6   & 75.6$\pm$3.9    & 85.3$\pm$4.7    & 93.3$\pm$1.8   & 81.7$\pm$5.3   & 84.0  & ~86.3 \Tstrut \\
			Joint+KD                              & 39.32M  & 84.6$\pm$2.6  & 89.0$\pm$3.3           & 89.8$\pm$4.1   & 92.5$\pm$3.7   & 89.0   & 76.6$\pm$4.2    & 85.2$\pm$6.7    & 93.2$\pm$2.0   & 83.0$\pm$3.0   & 84.5   & 86.8 \\
			``Y"-shaped~\cite{nie2016fully}       & 39.99M  & 86.7$\pm$4.8  & 90.9$\pm$5.3           & 92.2$\pm$2.5   & 87.4$\pm$5.1   & 89.3   & 78.6$\pm$2.6    & 84.2$\pm$9.1    & 93.0$\pm$2.0   & 81.2$\pm$2.1   & 84.3  & 86.8  \\
			``X"-shaped~\cite{valindria2018multi} & 65.96M  & 88.2$\pm$4.9  & 90.6$\pm$3.1           & 92.9$\pm$2.6   & 86.2$\pm$5.3   & 89.5   & 78.2$\pm$3.1    & 84.9$\pm$7.1    & 93.0$\pm$1.3   & 82.8$\pm$4.4   & 84.7   & 87.1 \\
			``Chilopod"-shaped                    & 39.34M  & 88.2$\pm$3.1  & 90.3$\pm$3.1           & 91.4$\pm$2.8   & 92.2$\pm$5.8   & 90.5   & 79.3$\pm$5.1    & 85.8$\pm$5.3    & 93.1$\pm$2.3   & 80.7$\pm$4.7   & 84.7   & 87.6 \\
			\textbf{Ours}                         & 39.34M  & \textbf{88.5$\pm$3.1}  & 91.5$\pm$3.1  & \textbf{93.1$\pm$2.1}  & \textbf{93.6$\pm$4.3}  & \textbf{91.7}  & \textbf{80.8$\pm$3.0}  & \textbf{86.5$\pm$6.5}  & \textbf{93.6$\pm$1.8}  & 83.1$\pm$5.8  & \textbf{86.0}  & \textbf{88.8} \\
			\hline
			& & \multicolumn{10}{c|}{\textbf{Hausdorff Distance (avg.$\pm$std., $mm$})} &  \Tstrut \Bstrut \\ 
			\hline
			Payer et al.~\cite{payer2017multi}    &  -       & - & - & - & - &  -  & - & - & - & - & - & - \Tstrut \\
			Individual   & 78.64M  &3.70$\pm$2.86&4.41$\pm$2.22&3.80$\pm$3.08 &3.56$\pm$3.43 &3.87 &1.87$\pm$0.99 &2.91$\pm$3.59 & 1.66$\pm$0.62 &2.48$\pm$2.47 &2.23 &3.05                 \\
			\hline
			Joint                                 & 39.32M  & 3.97$\pm$2.96  & 5.81$\pm$4.31  & 3.49$\pm$2.28   & 2.40$\pm$1.74   & 3.92   & 2.07$\pm$1.27    & 2.31$\pm$2.12    & 1.42$\pm$0.73   & 2.04$\pm$1.80   &\textbf{1.96}   & ~2.94 \Tstrut \\
			Joint+KD                              & 39.32M  & 3.05$\pm$1.71  & 4.42$\pm$4.94  & 2.77$\pm$2.08   & 1.88$\pm$1.75   & 3.03   & 1.97$\pm$1.30    & \textbf{2.29$\pm$2.19}    & 1.36$\pm$0.91   & 2.66$\pm$2.49   & 2.07   & 2.55 \\
			``Y"-shaped~\cite{nie2016fully}       & 39.99M  & 2.60$\pm$1.64  & 3.57$\pm$2.48  & 2.51$\pm$1.74  & 2.97$\pm$6.03  &2.91 &2.50$\pm$1.89  & 4.39$\pm$3.44  & 1.75$\pm$0.83  & 1.99$\pm$1.21  & 2.66 & 2.78  \\
			``X"-shaped~\cite{valindria2018multi} & 65.96M   & 2.61$\pm$1.60  & 3.46$\pm$3.13  & 2.37$\pm$1.63   & 2.84$\pm$7.32   & 2.82   & 2.49$\pm$1.56    & 3.86$\pm$2.54    & 2.19$\pm$5.10   & \textbf{1.97$\pm$1.40}   & 2.63   & 2.72 \\
			``Chilopod"-shaped                    & 39.34M  & \textbf{2.32$\pm$1.21}  & 3.08$\pm$2.37  & 2.32$\pm$2.10   & 2.48$\pm$3.32   & 2.55   & 2.30$\pm$1.34    & 2.82$\pm$3.17    & 1.38$\pm$0.63   & 3.75$\pm$3.93   & 2.56   & 2.56 \\
			\textbf{Ours}                         & 39.34M  & 2.38$\pm$1.72  & \textbf{2.43$\pm$1.70}  & \textbf{2.12$\pm$2.01}  & \textbf{1.74$\pm$1.91}  & \textbf{2.17}  & \textbf{1.85$\pm$1.21}  & 2.59$\pm$3.24  & \textbf{1.36$\pm$0.78}  & 2.30$\pm$2.31  & 2.02  & \textbf{2.10} \\
			\bottomrule
		\end{tabular}
	}
	\label{tab:table-cardiac}
	\vspace{-1mm}
\end{table*}

\begin{figure*}[t]
	\centering
	\includegraphics[width=0.96\textwidth]{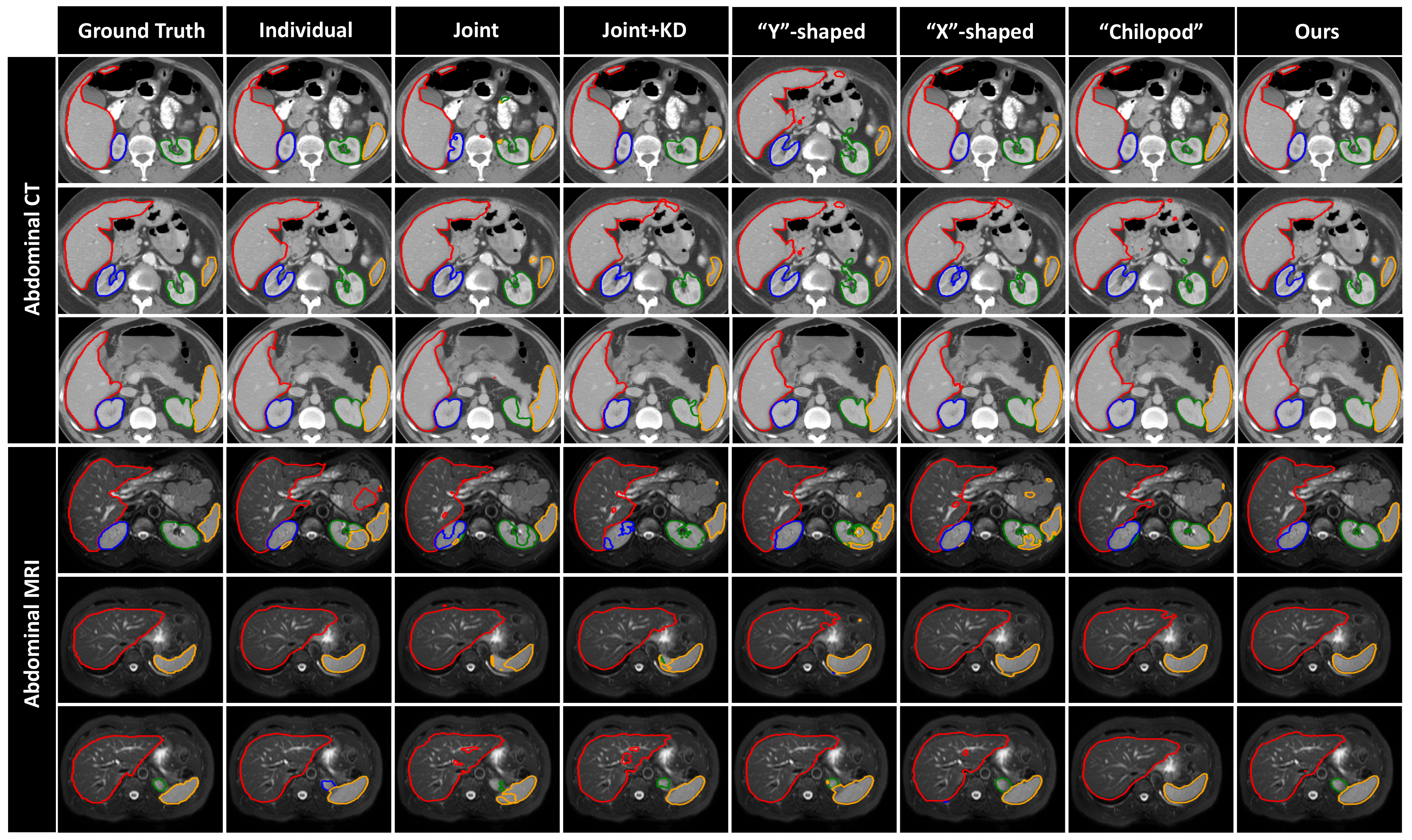}
	\vspace{-2mm}
	\caption{Qualitative comparison of the segmentation results from the seven different settings on abdominal multi-organ segmentation. The red, yellow, blue and green colors denote the boundary of liver, spleen, R-kdy and L-kdy (in human-body view), respectively.}
	\label{fig:fig-organ}
\end{figure*}

\begin{table*}[h!]
	\centering
	\caption{\small{Multi-modal learning results of abdominal multi-organ segmentation with 3D network under different settings.}}
	\vspace{-2mm}
	\scalebox{0.85}{
		\begin{tabular}{c|c|ccccc|ccccc|cc}
			\toprule
			\multirow{2}{*}{Methods}        &  Param  & \multicolumn{5}{c|}{Abdominal CT} & \multicolumn{5}{c|}{Abdominal MRI} & Overall \Tstrut \Bstrut \\ 
			\cline{3-12}
			& Scale   & Liver   & Spleen & R-kdy & L-kdy  & Mean  & Liver  & Spleen  & R-kdy & L-kdy & ~Mean  & ~Mean \Tstrut \\
			\hline
			& & \multicolumn{10}{c|}{\textbf{Dice Coefficient (mean$\pm$std, $\%$)}} &  \Tstrut \Bstrut \\ 
			\hline
			Individual    & 3832K   &\textbf{93.1$\pm$1.3}    & 91.6$\pm$2.4  & 92.4$\pm$1.5  & 85.0$\pm$5.6    & 90.5  & 87.0$\pm$3.3   & 74.0$\pm$3.3    & 90.9$\pm$2.0  & 83.0$\pm$6.5  & 83.7  & 87.1\Tstrut \\
			\hline
			Joint           & 1916K   & 89.4$\pm$3.4    & 88.4$\pm$1.4  & 72.3$\pm$4.7  & 74.5$\pm$12.5    & 81.2  & 81.9$\pm$1.8   & 63.7$\pm$6.8    & 78.0$\pm$0.1  & 82.9$\pm$4.0  & 76.7  & 79.0\Tstrut \\
			Joint+KD        & 1916K   &84.7$\pm$3.1    & 83.4$\pm$3.4  & 88.4$\pm$2.0  & 82.8$\pm$5.1    & 84.8  & 83.1$\pm$3.3   & 62.4$\pm$8.4    & 88.0$\pm$0.9  & 74.9$\pm$6.2  & 77.1  & 81.0\\
			``Y"-shaped~\cite{nie2016fully}     & 2124K   & 90.5$\pm$2.4  & 92.6$\pm$1.4  & 92.0$\pm$2.4  &84.8$\pm$8.2  & 90.0  & 89.3$\pm$3.1   & 82.4$\pm$2.5    & 89.6$\pm$0.6  & 80.8$\pm$8.9  & 85.5  & 87.8\\
			``X"-shaped~\cite{valindria2018multi}     & 3389K   & 92.3$\pm$1.2    & 92.9$\pm$0.9  & 90.9$\pm$2.9  & 84.4$\pm$6.7    & 90.1  & 88.5$\pm$3.0   & 84.8$\pm$1.4    & 89.2$\pm$0.1  & 86.0$\pm$4.1  & 87.1  & 88.7\\
			``Chilopod"-shaped     & 1919K   & 91.5$\pm$1.7    & 93.0$\pm$1.6  & 92.2$\pm$2.1  & 87.4$\pm$4.7    & 91.0  & 89.8$\pm$1.5   & 81.7$\pm$6.7    & \textbf{91.1$\pm$1.7}  & 88.2$\pm$2.5  & 87.7    & 89.3\\
			\textbf{Ours}   & 1919K   & 92.7$\pm$1.8  & \textbf{93.7$\pm$1.7}  & \textbf{94.0$\pm$0.7}  & \textbf{89.5$\pm$3.9}  & \textbf{92.4}  & \textbf{90.3$\pm$2.8}  & \textbf{87.4$\pm$1.1}  & 91.0$\pm$1.5 & \textbf{88.3$\pm$1.7}     & \textbf{89.3}  & \textbf{90.8} \\
			\hline
			& & \multicolumn{10}{c|}{\textbf{Hausdorff Distance (mean$\pm$std, $mm$)}} &  \Tstrut \Bstrut \\ 
			\hline
			Individual      & 3832K   & \textbf{3.08$\pm$4.27}    & 1.99$\pm$2.29  & 1.31$\pm$0.88  & 2.96$\pm$2.98    & 2.34  & 4.11$\pm$1.87   & 2.54$\pm$1.19    & 1.49$\pm$0.63  & 4.45$\pm$1.21  &3.15  & 2.74\Tstrut \\
			\hline
			Joint           & 1916K   & 4.46$\pm$7.34    & 2.01$\pm$1.81  & 2.93$\pm$1.49  & 2.76$\pm$2.61    & 3.04  & 4.77$\pm$1.82   & 5.88$\pm$6.18    & 2.84$\pm$1.12  & 5.08$\pm$5.56  & 4.65  & 3.85\Tstrut \\
			Joint+KD        & 1916K   & 4.71$\pm$3.23    & 2.54$\pm$2.07  & 2.08$\pm$1.08  & 2.51$\pm$1.71    & 2.96  & 5.04$\pm$2.43   & 3.53$\pm$1.62    & 1.97$\pm$0.75  & 5.58$\pm$9.82  & 4.03  & 3.50\\
			``Y"-shaped~\cite{nie2016fully}     & 2124K   & 3.75$\pm$4.43    & 1.59$\pm$2.72  & 1.40$\pm$0.86  & 2.59$\pm$2.90 &2.45   & 3.95$\pm$4.17  & 1.84$\pm$0.66   & 1.35$\pm$0.51    & 4.06$\pm$1.14  & 2.80  & 2.57\\
			``X"-shaped~\cite{valindria2018multi}     & 3389K   & 3.50$\pm$4.84    & 1.67$\pm$1.83  & 1.51$\pm$1.01  & 2.45$\pm$2.18    & 2.29  & 4.67$\pm$2.96   & 2.45$\pm$2.69    & 1.56$\pm$0.98  & \textbf{3.85$\pm$0.68}  & 3.14  & 2.71\\
			``Chilopod"-shaped     & 1919K   & 3.68$\pm$3.92    & \textbf{1.53$\pm$1.63}  & 1.32$\pm$0.72 & 1.93$\pm$1.53    & 2.12  & 3.79$\pm$4.68   & 1.85$\pm$0.83    & \textbf{1.29$\pm$0.51}  & 4.33$\pm$4.08  & 2.81  & 2.47\\
			\textbf{Ours}   & 1919K   & 3.27$\pm$3.74  & 1.59$\pm$2.09  & \textbf{1.20$\pm$0.81}  & \textbf{1.86$\pm$1.57}  & \textbf{1.98}  & \textbf{3.36$\pm$3.15}  & \textbf{1.82$\pm$0.60}  & 1.38$\pm$0.42 & 4.27$\pm$3.40     & \textbf{2.71}  & \textbf{2.34} \\
			\bottomrule
		\end{tabular}
	}
	\label{tab:table-organ}
\end{table*}

\subsection{Experimental settings}

We generally use BN layer, as it is the most widely-adopted normalization technique in medical image segmentation tasks.
Ablation study on other normalization layers is also conducted.
For comprehensive analysis and comparison, we design the following seven experimental settings, and implement on our datasets.
Network architecture and hyper-parameters are fixed for all the settings, for a fair comparison of different methods.
\\
\vspace{-6mm}
\\
\begin{itemize}
	\item \textbf{Individual}: independently training a separate model for each individual modality.
	\item \textbf{Joint}: training a single model with all network parameters (convolution kernels and BN) shared for CT and MRI.
	\item \textbf{Joint+KD}: training a joint model for CT and MRI, with adding our proposed KD-loss.
	\item \textbf{``Y"-shaped}~\cite{nie2016fully}: modality-specific encoders and shared decoders, which is a widely-used late fusion scheme for multi-modal learning.
	\item \textbf{``X"-shaped}~\cite{valindria2018multi}: modality-specific encoder and decoder layers, with shared bottleneck layers. The~\cite{valindria2018multi} demonstrate this two-stream architecture to be the current state-of-the-art for unpaired CT and MRI segmentation.
	\item \textbf{``Chilopod"-shaped}: our proposed architecture, i.e., sharing all CNN kernels while keeping internal feature normalization layers as modality-specific.
	\item \textbf{Ours}: our full multi-modal learning scheme, i.e., using ``Chilopod"-shaped architecture and KD-loss together.
\end{itemize}

\subsection{Segmentation results and comparison with state-of-the-arts}
We evaluate the segmentation performance with 
the metrics of volume Dice coefficient (\%) and surface Hausdorff distance ($mm$) by calculating the average and standard deviation of the segmentation results for each class,
as listed in Table~\ref{tab:table-cardiac} and Table~\ref{tab:table-organ} respectively for the two different tasks.
The mean Dice coefficient and Hausdorff distance over each modality as well as over two modalities are also presented for a straight-forward comparison.
Our implemented \textit{Individual} models are the baselines from single modality training. We compare the performance of different multi-modal learning methods, including two state-of-the-art approaches~\cite{nie2016fully,valindria2018multi}. We also refer to the available winning performance of the challenge~\cite{payer2017multi,zhuang2019evaluation} to demonstrate effectiveness of multi-modal learning.

\subsubsection{Results on multi-modal cardiac segmentation}
In Table~\ref{tab:table-cardiac}, we see that \textit{Joint} model obtains average segmentation Dice of 88.6\% on CT and 84.0\% on MRI, which are quite reasonable compared with \textit{Individual} model (88.4\% on CT and 84.8\% on MRI). This indicates that networks present sufficient capacity to analyze both CT and MRI in a compact model, though the data distributions of these two modalities are very different.
On top of \textit{Joint} model, adding our proposed KD-loss can improve the segmentation performance to 89.0\% on CT and 84.5\% on MRI, thank to the explicit guidance from confusion matrix alignment of the distilled semantic knowledge. Next, we compare the different methods which use modality-specific parameters for multi-modal learning.
We see that the models of \textit{``Y"-shaped}, \textit{``X"-shaped} and \textit{``Chilopod"} generally get higher performance over \textit{``Individual"} training.
The three models employ the same segmentation loss function (i.e., combining Dice loss and cross-entropy loss), with different ways of designing modality-specific and shared parameters for feature fusion. Both~\cite{nie2016fully} and \cite{van2019learning} utilize independent encoders/decoders for each modality, while we just use modality-specific BN layers resulting in a more compact model. By further leveraging KD-loss as sort of cross-modality transductive bias, the segmentation performance is boosted to overall Dice of 88.8\% (specifically, 91.7\% on CT and 86.0\% on MRI), exceeding our own implemented \textit{``Indivudial"} training as well as the MICCAI-MMWHS challenge winner Payer et al.~\cite{payer2017multi} (overall Dice of 85.5\%) which used single model learning.
Our approach also achieves the lowest overall mean Hausdorff distance (i.e., $2.10mm$) among all the compared methods.
Fig.~\ref{fig:fig-cardiac} presents typical segmentation results of CT and MRI images, for a quantitative comparison of the different methods.

\subsubsection{Results on multi-modal abdominal organ segmentation}
Table~\ref{tab:table-organ} lists results of multi-organ segmentation using 3D model with skip connections. When analyzing CT and MRI in a \textit{Joint} model, the performance shows a large decrease compared with \textit{Individual} training,
i.e., overall average Dice dropping from 87.1\% to 79.0\% while mean Hausdorff distance increases from 2.74$mm$ to 3.85$mm$. This indicates that the multi-modal shift may present more challenges when learning in 3D.
Adding our KD-loss to guide the convergence towards extracting universally representative features, the segmentation Dice is improved by 2.0\% (from 79.0\% to 81.0\%).
For multi-modal learning methods, We observe that \textit{``X"-shaped} model is superior to \textit{``Y"-shaped} model, which is consistent with the findings in~\cite{valindria2018multi}.
Our proposed \textit{``Chilopod"-shaped} model, 
i.e., sharing all the convolution kernels but with modality-specific BN layers, achieves comparable or better performance than the \textit{``X"-shaped} model, with a higher parameter efficiency.
Further adding our KD-loss, our full multi-modal learning scheme achieves the best performance on average segmentation Dice of 90.8\% and average Hausdorff distance of 2.34$mm$, outperforming all \textit{Individual} and \textit{``X"-/``Y"-shaped} models by a large margin.
The Dice of challenge winners for the two datasets used in this multi-organ segmentation task are currently not available.
Last but not least, it is worth noting that, \textit{Ours} significantly improves the segmentation performance on MRI over \textit{Individual} model by 5.6\% (from 83.7\% to 89.3\%).
This demonstrates that our method can effectively improve the performance on the modality with fewer training samples, by leveraging multi-modal learning. 
We notice that such improvement mainly comes from two organs: spleen and left kidney. These are located nearby and have similar appearance in MRI and pose challenges in the context of data scarcity. Multi-modal learning seems beneficial by enhancing high-level inter-class relationship alignment.
The Fig.~\ref{fig:fig-organ} presents typical segmentation results of CT and MRI images, for a quantitative comparison of these different methods.

\subsubsection{Statistical analysis on significance}
We have computed p-values using Student's t-tests when comparing our segmentation results with other methods, with numbers given in Table~\ref{tab:pvalue_cardiac} for 2D cardiac segmentation and Table~\ref{tab:pvalue_organ_w/o} for 3D abdominal organ segmentation. It is observed that we get $p<0.05$ in all settings, indicating a significant improvement for our approach. The statistical tests are conducted by jointly considering both results of CT and MRI in each setting.

\begin{table}[t]
	\centering
	\caption{\small{P-values for statistical analysis of all settings for our proposed method on 2D cardiac segmentation.}}
	\vspace{-2mm}
	\scalebox{0.68}{
		\begin{tabular}{c|cccccc}
			\toprule
			Metrics	& Individual & Joint &Joint+KD &``Y"-shaped~\cite{nie2016fully} &``X"-shaped~\cite{valindria2018multi} &``Chilopod"\\
			\hline
			Dice coefficient & 0.015 & 6e-4 &0.003 &0.006 &0.002 &0.029 \Tstrut\\
			
			Hausdorff distance &0.022 &0.036 &0.035 &0.044 &0.035 &7e-4\\
			\bottomrule
		\end{tabular}
	}
	\label{tab:pvalue_cardiac}
	\vspace{-2mm}
\end{table}

\begin{table}[t]
	\centering
	\caption{\small{P-values for statistical analysis of all settings for our proposed method on 3D organ segmentation.}}
	\vspace{-2mm}
	\scalebox{0.68}{
		\begin{tabular}{c|cccccc}
			\toprule
			Metrics & Individual & Joint &Joint+KD &``Y"-shaped~\cite{nie2016fully} &``X"-shaped~\cite{valindria2018multi} &``Chilopod"\\
			\hline
			Dice coefficient   & 0.009 & 4e-5 & 3e-4 & 0.015 & 0.005 & ~0.003 \Tstrut\\
			
			Hausdorff distance & 0.006 & 2e-9 & 5e-7 & 0.039 & 0.043 &0.036 \\
			\bottomrule
		\end{tabular}
	}
	\label{tab:pvalue_organ_w/o}
\end{table}

\subsubsection{Analysis on parameter efficiency}
A benefit of our multi-modal learning network is its high compactness.
With careful internal normalization of features, we can make more sufficient use of the remarkable capacity inherent in neural networks. We compute the parameter scales of all the models for our implemented seven different settings, as listed in the second column of Table~\ref{tab:table-cardiac} and Table~\ref{tab:table-organ}. A \textit{Joint} model has 39.32M parameters for 2D model and 1916K parameters for 3D model.
Our 2D model has more parameters because it is much deeper and wider than the 3D model. 
With \textit{individual} training, we need double of the parameter scales (i.e., 78.64M for 2D and 3832K for 3D), since each single modality has its own model separately. Using \textit{Joint+KD} adds no extra parameters, while the distilled knowledge helps stimulate underlying cross-modality information.
The conventional \textit{``Y"-shaped} and \textit{``X"-shaped} multi-modal learning schemes consist more parameters due to their modality-specific encoders/decoders. Specifically, the \textit{``X"-shaped} model almost doubles the parameter scale, for the cost of using more modality-specific layers.
In comparison, our proposed multi-modal learning schemes (\textit{``Chilopod"} and \textit{Ours}) only need marginally extra parameters for separate internal feature normalization with BN (parameterized by a set of feature channel tied scalars $\{\gamma, \beta\}$. Specifically, our method only adds 0.02M for 2D network and 3K parameters for 3D network, which is quite parameter efficient compared with existing multi-modal schemes.

\subsection{Analytical ablation studies}

\begin{figure}[t]
	\centering
	\includegraphics[width=0.4\textwidth]{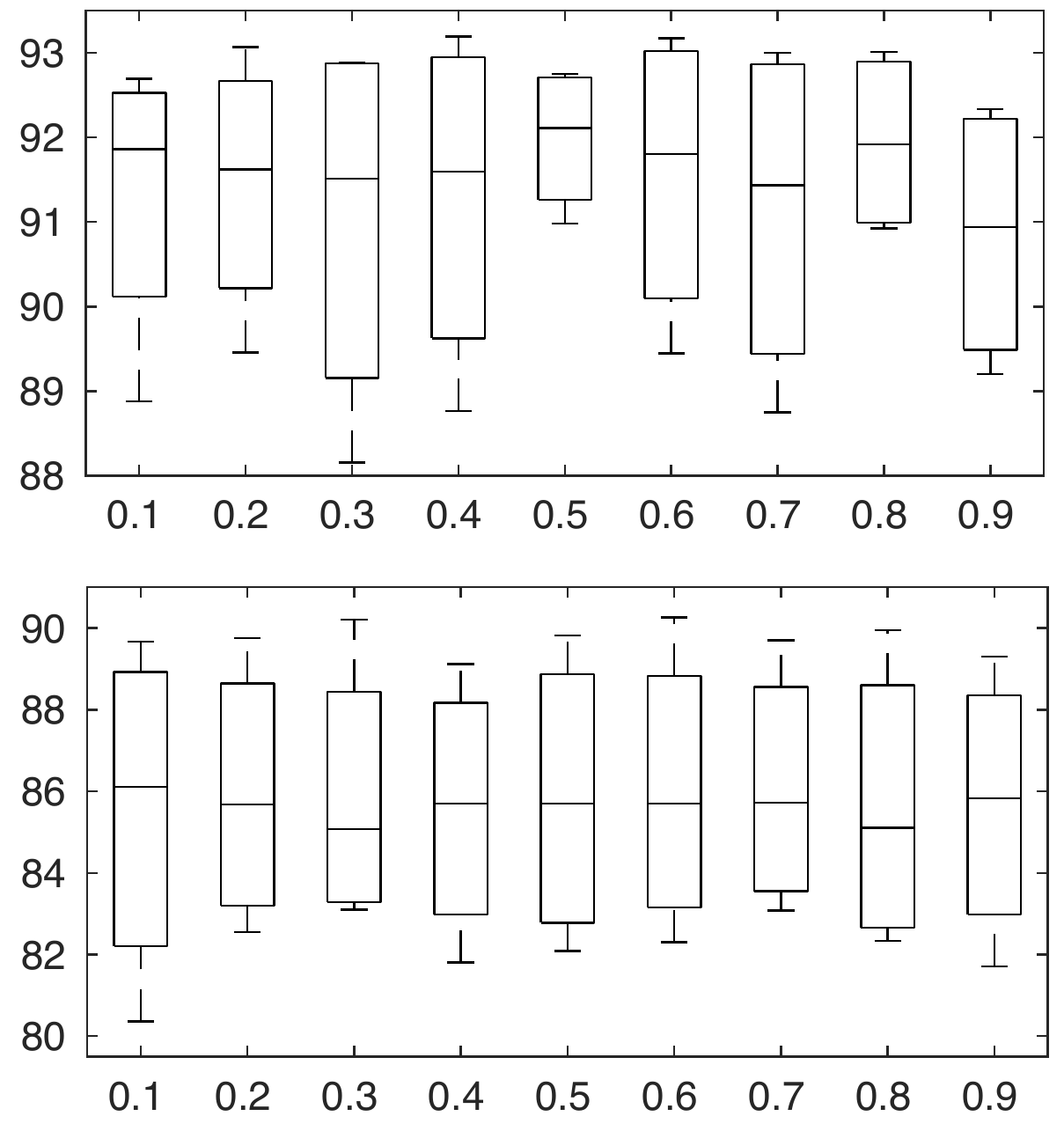}
	\vspace{-3mm}
	\caption{Box-plots of Dice (average over all structures) on CT (top) and MRI (bottom) for cardiac segmentation, when varying $\alpha$ from 0.1 to 0.9.}
	\label{fig:box-plot}
\end{figure}

\subsubsection{Different internal normalization layers}
We demonstrate that the effectiveness of separate internal feature normalization is agnostic to different ways of grouping features for $S_g^k$. Three additional popular feature normalization methods (i.e., Instance Norm~\cite{ulyanov2016instance}, Layer Norm~\cite{ba2016layer}, and Group Norm~\cite{wu2018group}) are implemented for five ablation settings of our method, using the cardiac segmentation dataset. The results are listed in Table~\ref{tab:results-norm}. 
\textit{Ours} consistently presents superior performance over \textit{Individual} learning for all popular feature normalization methods.
Comparing the results of \textit{``Joint"} models under these different normalization methods, we notice that Instance Norm has the best average Dice, achieving 87.5\%.
This indicates that more refined internal feature normalization benefits multi-modal learning under great parameter sharing.
This also meets our hypothesis on normalizing multi-modal images separately in our compact model.

\subsubsection{Different weights of KD-loss}
We vary the trade-off weight of our KD-loss, to analyze its sensitivity to the hyper-parameter of $\alpha$ in Eq.~\ref{eq:overall_loss}.
Specifically, we range $\alpha \in [0.1,0.9]$ at a step of 0.1, and observe our multi-modal segmentation performance on the cardiac dataset.
In Fig.~\ref{fig:box-plot}, the box-plots present the mean of segmentation Dice across all classes on CT (top) and MRI (bottom).
Note that the setting of $\alpha \! = \!0$ corresponds to the seventh and twelfth columns in Table~\ref{tab:table-cardiac}.
We observe that our method can generally improve the segmentation performance over baselines (e.g., 88.4\% for CT and 84.8\% for MRI at \textit{``Individual"} training), while not being very sensitive to the value of $\alpha$.

\begin{table*}[t]
	\centering
	\caption{Multi-modal cardiac segmentation with a 2D network by using different feature normalization layers (Dice, \%).}
	\scalebox{0.9}{
		\begin{tabular}{c|c|ccccc|ccccc|c}
			\toprule
			\multirow{2}{*}{Normalization} & \multirow{2}{*}{Methods} & \multicolumn{5}{c|}{Cardiac CT} & \multicolumn{5}{c|}{Cardiac MRI} & Overall\\ 
			\cline{3-12}
			Type & &  LVM & LAC & LVC  &  AA & Mean   & LVM & LAC & LVC  &  AA & Mean & Mean \Tstrut  \\
			\hline
			& Individual  & 87.5$\pm$3.2 & 90.2$\pm$3.8 & 92.1$\pm$2.4 & ~87.4$\pm$14.7& 89.3 & 80.8$\pm$3.3 & 86.6$\pm$5.4 & 92.7$\pm$1.7 & 82.6$\pm$5.6 & 85.7 & ~87.5 \Tstrut  \\
			Instance        & Joint       & 86.6$\pm$2.4 & 88.7$\pm$2.9 & 91.3$\pm$2.9 & 92.7$\pm$4.6 & 89.9 & 82.1$\pm$3.1 & 83.7$\pm$8.7 & 93.9$\pm$2.0 & 80.7$\pm$6.6 & 85.1 & 87.5 \\
			Normalization   & Joint+KD    & 86.7$\pm$1.8 & 90.8$\pm$3.7 & 91.7$\pm$3.1 & 92.4$\pm$6.6 & 90.4 & 80.9$\pm$4.3 & 85.4$\pm$7.3 & 93.9$\pm$2.1 & 82.0$\pm$7.4 & 85.5 & 88.0 \\
			& "Chilopod"  & 86.5$\pm$2.6 & 91.5$\pm$3.0 & 91.7$\pm$3.2 & 94.2$\pm$4.3 & 91.0 & 81.2$\pm$4.7 & 85.4$\pm$6.8 & 94.1$\pm$1.7 & 81.0$\pm$3.7 & 85.4 & 88.2 \\
			& Ours        & 87.3$\pm$2.1 & 90.9$\pm$2.8 & 92.6$\pm$3.0 & 94.0$\pm$4.0 & 91.2 & 82.4$\pm$3.1 & 85.6$\pm$5.6 & 94.1$\pm$2.0 & 82.6$\pm$5.5 & 86.2  & 88.7 \\
			\hline
			& Individual  & 87.8$\pm$3.9 & 89.9$\pm$2.7 & 92.7$\pm$2.2 & 90.2$\pm$9.2 & 90.2 & 79.2$\pm$4.3 & 86.2$\pm$4.5 & 91.4$\pm$2.9 & 83.9$\pm$3.1 & 85.2 & ~87.7 \Tstrut  \\
			Layer           & Joint       & 83.0$\pm$4.5 & 88.2$\pm$2.0 & 90.2$\pm$2.9 & 92.2$\pm$1.5 & 88.4 & 76.6$\pm$3.5 & 83.3$\pm$6.6 & 92.6$\pm$2.1 & 77.3$\pm$6.4 & 82.5 & 85.4  \\
			Normalization   & Joint+KD    & 86.1$\pm$3.0 & 90.7$\pm$3.6 & 92.4$\pm$3.6 & 94.6$\pm$4.5 & 90.9 & 79.1$\pm$3.9 & 85.8$\pm$6.7 & 92.3$\pm$1.9 & 81.6$\pm$5.7 & 84.7 & 87.8  \\
			& "Chilopod"  & 86.6$\pm$1.9 & 90.9$\pm$3.2 & 91.5$\pm$2.9 & 94.2$\pm$1.5 & 90.8 & 78.2$\pm$3.4 & 85.0$\pm$7.1 & 92.7$\pm$1.8 & 80.3$\pm$2.7 & 84.1 & 87.4  \\
			& Ours        & 85.8$\pm$4.0 & 90.6$\pm$3.1 & 92.3$\pm$2.5 & 94.3$\pm$2.7 & 90.7 & 79.4$\pm$4.2 & 86.2$\pm$6.3 & 92.5$\pm$2.0 & 82.0$\pm$5.7 & 85.0 & 87.9 \\
			\hline
			& Individual  & 88.1$\pm$3.7 & 89.8$\pm$3.0 & 92.4$\pm$2.1 & ~84.1$\pm$10.5& 88.6 & 79.0$\pm$3.8 & 84.5$\pm$5.5 & 91.5$\pm$4.6 & 82.0$\pm$5.4 & 84.2 & ~86.4  \Tstrut \\
			Group           & Joint       & 86.8$\pm$3.0 & 89.5$\pm$2.7 & 91.6$\pm$2.6 & 91.0$\pm$3.8 & 89.7 & 79.7$\pm$3.2 & 85.0$\pm$5.5 & 92.7$\pm$2.2 & 79.0$\pm$6.2 & 84.1 & 86.9  \\
			Normalization   & Joint+KD    & 85.9$\pm$2.6 & 90.3$\pm$3.5 & 91.6$\pm$3.5 & 93.6$\pm$4.2 & 90.4 & 80.8$\pm$4.1 & 86.8$\pm$6.2 & 93.4$\pm$2.1 & 82.0$\pm$5.2 & 85.7 & 88.0 \\
			& "Chilopod"  & 86.1$\pm$4.7 & 91.8$\pm$2.0 & 91.8$\pm$4.2 & 93.6$\pm$3.0 & 90.8 & 78.5$\pm$5.2 & 85.2$\pm$7.6 & 92.5$\pm$1.3 & 79.1$\pm$6.9 & 83.8 & 87.3 \\
			& Ours        & 88.4$\pm$2.8 & 90.4$\pm$3.1 & 92.3$\pm$2.4 & 93.7$\pm$4.1 & 91.2 & 81.8$\pm$1.5 & 85.0$\pm$6.8 & 93.9$\pm$1.5 & 81.9$\pm$4.2 & 85.6 & 88.4 \\
			\bottomrule
		\end{tabular}
	}
	\label{tab:results-norm}
\end{table*}

\subsubsection{Learning curve of KD-loss}
\begin{figure}[t]
	\centering
	\includegraphics[width=0.45\textwidth]{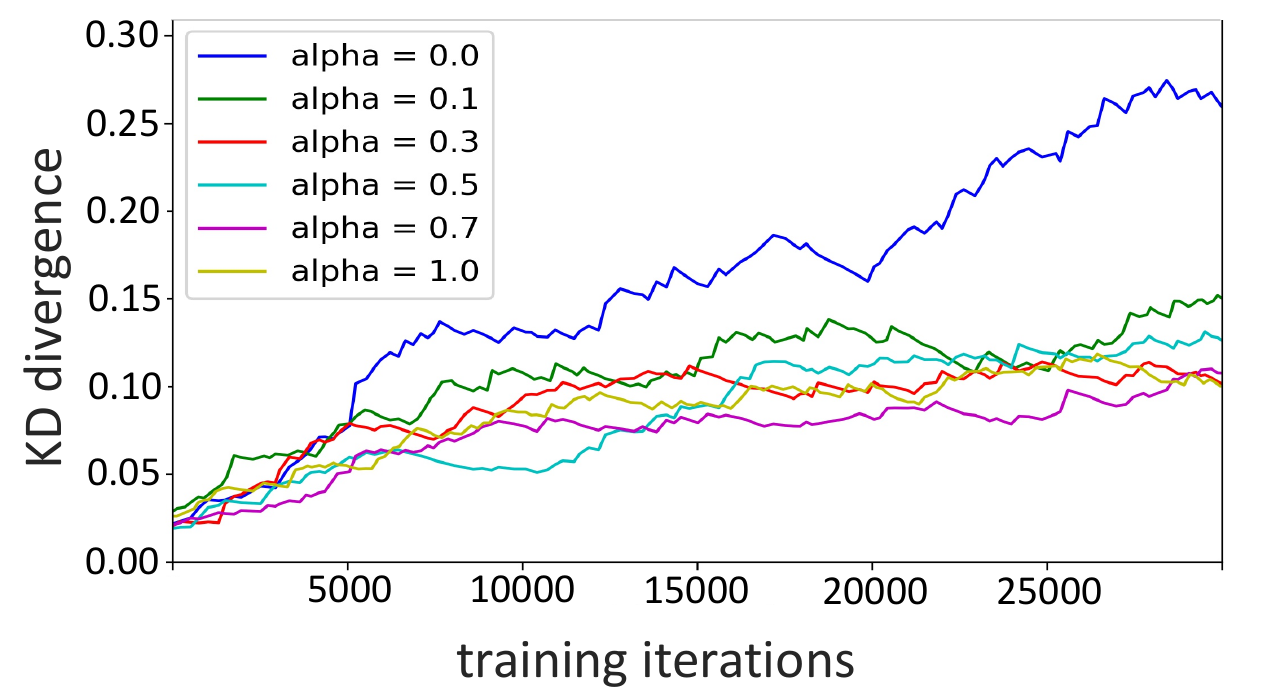}
	\vspace{-3mm}
	\caption{Comparison of learning curves for KD-loss at different values of $\alpha$.}
	\label{fig:curve}
\end{figure}

In Fig.~\ref{fig:curve}, we present the learning curve of KD-loss computed on test data at multi-organ segmentation dataset. We observe that without constraints (i.e., $\alpha=0$), the distilled knowledge from two modalities diverges. By activating KD-loss, the $\mathcal{L}_{\text{kd}}$ is stabilized, reflecting that the probability distributions across classes are better aligned. This observation also explains the performance gain from using the KD-loss as guidance of high-level semantic alignment.

\subsubsection{Evolution of prediction alignment between modalities}
\begin{figure}[t]
	\centering
	\vspace{-2mm}
	\includegraphics[width=0.48\textwidth]{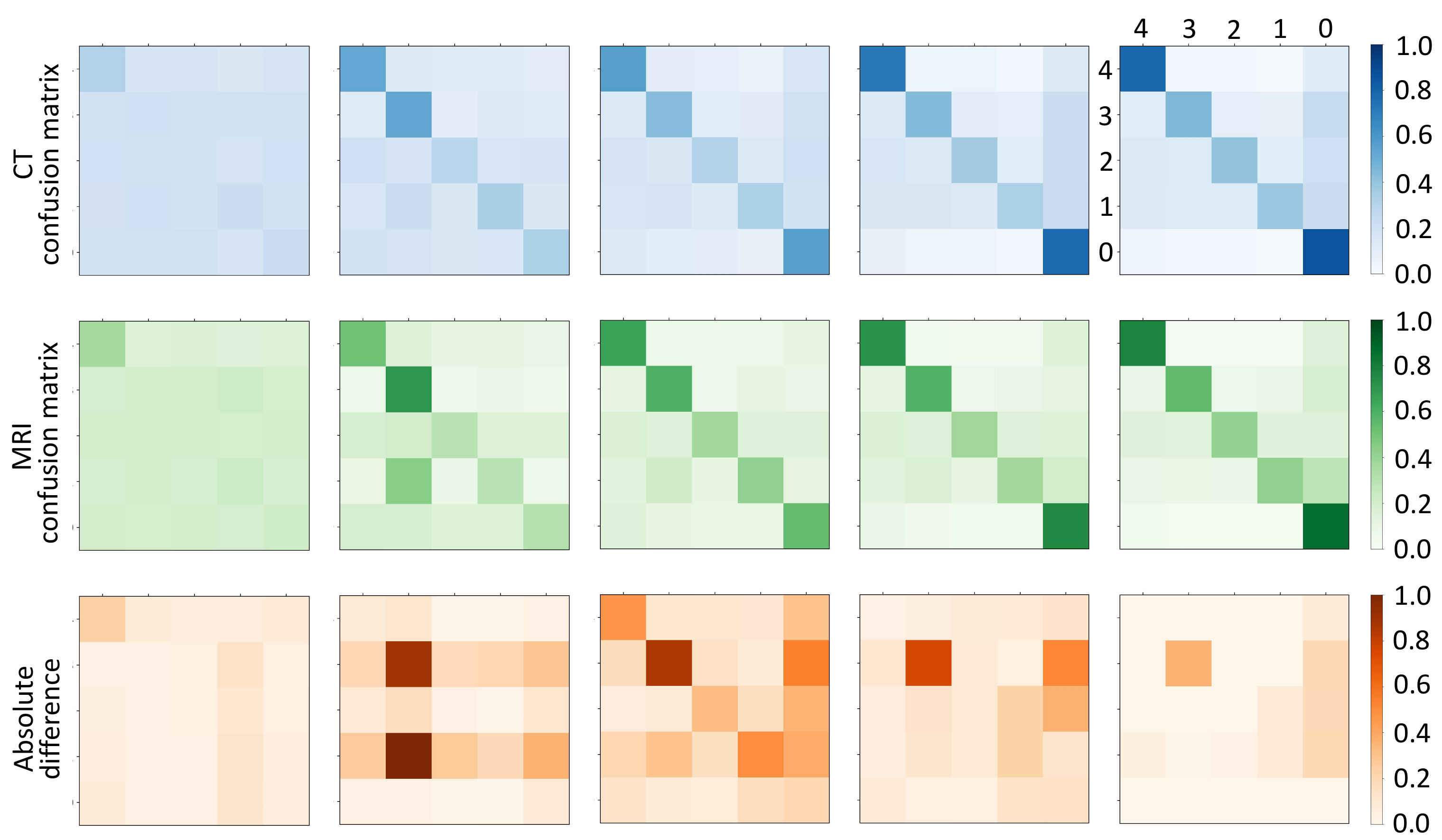}
	\vspace{-3mm}
	\caption{Visualization of the evolution of the alignment of prediction confusion matrices between CT (first row) and MRI (second row) modalities as training going on. For better illustration of differences, the third row shows their absolute difference planes (with values amplified by a five-times scaling). The classes of $4,3,2,1,0$ respectively correspond to liver, spleen, right kidney and left kidney in our 3D multi-organ segmentation task.}
	\label{fig:confusion}
\end{figure}

In Fig.~\ref{fig:confusion}, we visualize the evolution of the confusion matrices (i.e., $\mathbf{q}^a$ and $\mathbf{q}^b$) for both modalities, from the beginning of training until model convergence. To more clearly observe alignment of these matrices (i.e., Eq.~4), we compute their absolute difference plane by abstracting one from the other, as illustrated in the bottom row. It is observed that when the model is randomly initialized, the confusion matrices are invariable with no much difference between CT and MRI. As training goes on, the model starts learning and gradually produces meaningful confusion planes with stronger diagonal responses. Notably, with the model converging, the difference plane between CT and MRI returns to clean again, but for the reason of successful alignment of their confusion matrices.

\section{Discussion}

This work tackles the challenging task of multi-class segmentation on unpaired CT and MRI images. The difficulties mainly arise from the significant distribution shift and absence of registration between the two modalities.
This problem has not been well studied so far, compared with segmentation on paired multi-sequence of MRI images. Valindria et al.~\cite{valindria2018multi} is the current state-of-the-art study on this topic, with empirical results demonstrating that leveraging cross-modality information is superior to single model learning. 
To more explicitly utilize cross-modality knowledge, we present a novel approach by distilling the semantic representations in a high-level within a compact model. Notably, our multi-modal learning scheme is architecture-independent, therefore can be easily integrated into various existing 2D and 3D CNN models.
Our approach is developed for unpaired multi-modal learning, and it does not make use of any information which relies on image alignment. In this regard, we think that matching the distributions of the predictions on \emph{aligned images} would not bring much extra help to our approach as it is. Nevertheless, if such aligned images are available, our method is still directly applicable, and could be extended to leverage the fine-grained pixel-wise alignment.

The remarkable capacity of deep neural networks motivates our design of a highly compact model for multi-modal learning. 
Bilen et al.~\cite{bilen2017universal} show that sharing all kernels (even including the final classifier) while using domain-specific batch normalization can work reasonably well for different easy tasks, e.g., MNIST and CIFAR-10.
In medical imaging, Moeskops et al.~\cite{moeskops2016deep} build a single network for different segmentation tasks in different modalities.
Karani et al.~\cite{karani2018lifelong} successfully adapt an MRI brain segmentation model to different scanners/protocols by only fine-tuning the BN layers.
Inspired by these findings, we argue that a single shared network has the potential to work well on very different multi-modal data with similar structures (e.g., CT and MRI) requiring only a few modality-specific parameters.
Separate internal normalization may be sufficient to unleash the potential model capacity, not necessarily using a separate encoder for each modality.
An explicit regularization loss towards cross-modality semantic alignment helps further stimulate the model capacity.

\begin{figure}[t]
	\includegraphics[width=0.47\textwidth]{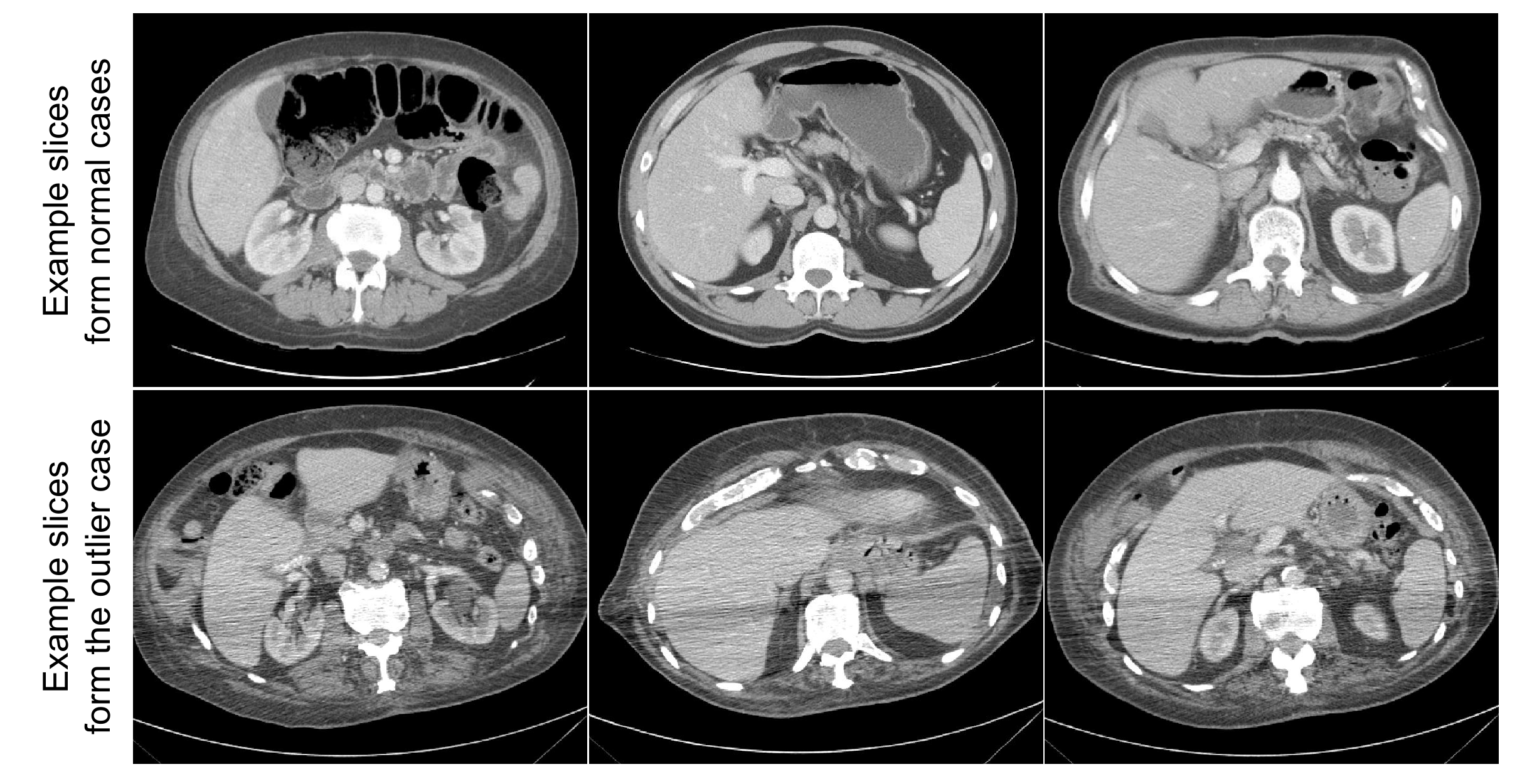}
	\vspace{-3mm}
	\caption{Illustration of example CT slices from general cases and an outlier case with artefacts, in the task of multi-organ segmentation in abdominal data.}
	\label{fig:outlier}
\end{figure}

\begin{table}[t]
	\centering
	\caption{\small{Mean results of the four organs in abdominal CT on the outlier case using different methods.}}
	\scalebox{0.62}{
		\begin{tabular}{p{2.2cm}|p{1.0cm}p{0.5cm}p{1.0cm}p{1.8cm}p{1.9cm}p{1.2cm}p{0.8cm}}
			\toprule
			~~~~~~ Metrics & Individual & Joint &Joint+KD &``Y"-shaped~\cite{nie2016fully} &``X"-shaped~\cite{valindria2018multi} &``Chilopod" & \textbf{Ours}\\
			\hline
			~ Dice coefficient & ~~~73.7 &74.4 & ~~~78.0 & ~~~~~~69.8 &  ~~~~~~76.6 &  ~~~~~76.4 & \textbf{81.1} \Tstrut \\
			
			Hausdorff distance & ~~~4.19 &4.93 & ~~~4.23 & ~~~~~~5.83 &  ~~~~~~5.32 &  ~~~~~4.89 & \textbf{3.78}\\
			\bottomrule
		\end{tabular}
	}
	\label{tab:table-outlier}
\end{table}

\revise{
	We have conducted case study regarding an outlier case with image artefacts in the abdominal CT dataset. As illustrated in Fig.~\ref{fig:outlier}, the outlier case presents clear artefacts with worse image quality compared with other general cases,
	noting that all the training cases are good-quality images without artefacts.
	The mean results of the four abdominal organs using seven different settings are listed in Table~\ref{tab:table-outlier}, where we see that all comparison methods obtained a lower performance on this one case (compared with general results in Table II). Our approach achieved a higher Dice score with a smaller Hausdorff distance compared with the other methods, demonstrating our superior robustness at lower image quality.
}

One limitation of this paper lies in the plain network architectures used for multi-modal segmentation. The employed 2D dilation network for cardiac segmentation and 3D U-Net for abdominal multi-organ segmentation are relatively basic, compared with more complicated network designs for multi-modal learning.
This is reasonable as we currently focus on studying separate feature normalization and knowledge distillation between modalities, but this may limit the segmentation accuracy. We plan to integrate our multi-modal learning scheme into more well-designed networks (e.g., consisting multi-scale feature fusion) in our future work, seeking more accurate segmentation of the images with multi-modal learning.
This extension is quite natural, thanks to the flexibility of our proposed multi-modal learning scheme.

The characteristics of modality differences may play an important role affecting the performance of multi-modal learning methods.
It would be interesting to explore further whether there are limitations on certain functional or statistical relationships between intensity distributions that cannot be handled by our proposed approach. The fact that it works for CT and MRI is encouraging that our approach may work for a large family of relationships. In preliminary synthetic experiments, we could also confirm that an anti-correlation between intensity (i.e., one modality is the inverse of the other) does not pose a problem.
Regarding differences in image resolution between modalities, we would expect this not to impact the method too much due to the statistical approach of distribution matching. However, this would need to be confirmed for more extreme cases where one of the modalities is of much lower resolution.

In general, our proposed method of separate internal feature normalization and knowledge distillation loss can be applied to many other situations, when we use data with a mixture of distributions or domains. For instance, it can be used for model learning by aggregating images acquired from different clinical sites in real world scenarios. 
The data from different sites can be separately normalized within a network, so that one can make better use of multiple data sources.
The knowledge distillation loss derived from semantic representations can be applied for domain adaptation problems by leveraging alignment of inter-class relationships. We will explore these extensions in future work.

\section{Conclusion}

We present a novel multi-modal learning scheme for unpaired CT and MRI segmentation, with high parameter efficiency and a new KD-loss term. Our method is general for multi-modal segmentation tasks, and we have demonstrated its effectiveness on two different tasks and both 2D and 3D network architectures.
Integrating analysis of multi-modal data into a single parameter efficient network helps to ease deployment and improve usability of the model in clinical practice. 
Moreover, our KD-loss encouraging robust features has a potential to tackle model generalization challenges in medical image segmentation applications.

\bibliographystyle{IEEEtran.bst}
\bibliography{reference}

\begin{thebibliography}{10}
\providecommand{\url}[1]{#1}
\csname url@samestyle\endcsname
\providecommand{\newblock}{\relax}
\providecommand{\bibinfo}[2]{#2}
\providecommand{\BIBentrySTDinterwordspacing}{\spaceskip=0pt\relax}
\providecommand{\BIBentryALTinterwordstretchfactor}{4}
\providecommand{\BIBentryALTinterwordspacing}{\spaceskip=\fontdimen2\font plus
\BIBentryALTinterwordstretchfactor\fontdimen3\font minus
  \fontdimen4\font\relax}
\providecommand{\BIBforeignlanguage}[2]{{%
\expandafter\ifx\csname l@#1\endcsname\relax
\typeout{** WARNING: IEEEtran.bst: No hyphenation pattern has been}%
\typeout{** loaded for the language `#1'. Using the pattern for}%
\typeout{** the default language instead.}%
\else
\language=\csname l@#1\endcsname
\fi
#2}}
\providecommand{\BIBdecl}{\relax}
\BIBdecl

\bibitem{nikolaou2011mri}
K.~Nikolaou, H.~Alkadhi, F.~Bamberg, S.~Leschka, and B.~J. Wintersperger, ``Mri
  and ct in the diagnosis of coronary artery disease: indications and
  applications,'' \emph{Insights into imaging}, vol.~2, no.~1, pp. 9--24, 2011.

\bibitem{karim2018algorithms}
R.~Karim, L.~E. Blake, J.~Inoue, Q.~Tao, S.~Jia, R.~J. Housden \emph{et~al.},
  ``Algorithms for left atrial wall segmentation and thickness--evaluation on
  an open-source ct and mri image database,'' \emph{Medical image analysis},
  vol.~50, pp. 36--53, 2018.

\bibitem{chen2018voxresnet}
H.~Chen, Q.~Dou, L.~Yu, J.~Qin, and P.-A. Heng, ``Voxresnet: Deep voxelwise
  residual networks for brain segmentation from 3d mr images,''
  \emph{NeuroImage}, vol. 170, pp. 446--455, 2018.

\bibitem{zhang2015deep}
W.~Zhang, R.~Li, H.~Deng, L.~Wang, W.~Lin \emph{et~al.}, ``Deep convolutional
  neural networks for multi-modality isointense infant brain image
  segmentation,'' \emph{NeuroImage}, vol. 108, pp. 214--224, 2015.

\bibitem{moeskops2016automatic}
P.~Moeskops, M.~A. Viergever, A.~M. Mendrik, L.~S. de~Vries, M.~J. Benders, and
  I.~I{\v{s}}gum, ``Automatic segmentation of mr brain images with a
  convolutional neural network,'' \emph{IEEE transactions on medical imaging},
  vol.~35, no.~5, pp. 1252--1261, 2016.

\bibitem{fidon2017scalable}
L.~Fidon, W.~Li, L.~C. Garcia-Peraza-Herrera, J.~Ekanayake, N.~Kitchen,
  S.~Ourselin \emph{et~al.}, ``Scalable multimodal convolutional networks for
  brain tumour segmentation,'' in \emph{MICCAI}, 2017, pp. 285--293.

\bibitem{kamnitsas2017efficient}
K.~Kamnitsas, C.~Ledig, V.~F. Newcombe, J.~P. Simpson, A.~D. Kane, D.~K. Menon
  \emph{et~al.}, ``Efficient multi-scale 3d cnn with fully connected crf for
  accurate brain lesion segmentation,'' \emph{Medical image analysis}, vol.~36,
  pp. 61--78, 2017.

\bibitem{valverde2017improving}
S.~Valverde, M.~Cabezas, E.~Roura, S.~Gonz{\'a}lez-Vill{\`a}, D.~Pareto, J.~C.
  Vilanova \emph{et~al.}, ``Improving automated multiple sclerosis lesion
  segmentation with a cascaded 3d convolutional neural network approach,''
  \emph{NeuroImage}, vol. 155, pp. 159--168, 2017.

\bibitem{nie2016fully}
D.~Nie, L.~Wang, Y.~Gao, and D.~Shen, ``Fully convolutional network for
  multi-modality isointense infant brain segmentation,'' in \emph{ISBI}, 2016,
  pp. 1342--1345.

\bibitem{li20183d}
X.~Li, Q.~Dou, H.~Chen, C.~W. Fu, X.~Qi, D.~L. Belav{\`y} \emph{et~al.}, ``3d
  multi-scale fcn with random modality voxel dropout learning for
  intervertebral disc localization and segmentation from multi-modality mr
  images,'' \emph{Medical image analysis}, vol.~45, pp. 41--54, 2018.

\bibitem{wang2014computer}
S.~Wang, K.~Burtt, B.~Turkbey, P.~Choyke, and R.~M. Summers, ``Computer
  aided-diagnosis of prostate cancer on multiparametric mri: a technical review
  of current research,'' \emph{BioMed research international}, vol. 2014, 2014.

\bibitem{dolz2018hyperdense}
J.~Dolz, K.~Gopinath, J.~Yuan, H.~Lombaert, C.~Desrosiers, and I.~B. Ayed,
  ``Hyperdense-net: A hyper-densely connected cnn for multi-modal image
  segmentation,'' \emph{IEEE transactions on medical imaging}, vol.~38, no.~5,
  pp. 1116--1126, 2018.

\bibitem{dolz2018ivd}
J.~Dolz, C.~Desrosiers, and I.~B. Ayed, ``Ivd-net: Intervertebral disc
  localization and segmentation in mri with a multi-modal unet,'' in
  \emph{International Workshop and Challenge on Computational Methods and
  Clinical Applications for Spine Imaging}.\hskip 1em plus 0.5em minus
  0.4em\relax Springer, 2018, pp. 130--143.

\bibitem{li2019mman}
J.~Li, Z.~L. Yu, Z.~Gu, H.~Liu, and Y.~Li, ``Mman: Multi-modality aggregation
  network for brain segmentation from mr images,'' \emph{Neurocomputing}, vol.
  358, pp. 10--19, 2019.

\bibitem{valindria2018multi}
V.~V. Valindria, N.~Pawlowski, M.~Rajchl, I.~Lavdas, A.~G. Rockall
  \emph{et~al.}, ``Multi-modal learning from unpaired images: Application to
  multi-organ segmentation in {CT} and {MRI},'' in \emph{WACV}, 2018, pp.
  547--556.

\bibitem{zhang2018translating}
Z.~Zhang, L.~Yang, and Y.~Zheng, ``Translating and segmenting multimodal
  medical volumes with cycle-and shape-consistency generative adversarial
  network,'' in \emph{Proceedings of the IEEE Conference on Computer Vision and
  Pattern Recognition}, 2018, pp. 9242--9251.

\bibitem{dou2019pnp}
Q.~Dou, C.~Ouyang, C.~Chen, H.~Chen, B.~Glocker \emph{et~al.}, ``Pnp-adanet:
  Plug-and-play adversarial domain adaptation network at unpaired
  cross-modality cardiac segmentation,'' \emph{IEEE Access}, vol.~7, pp.
  99\,065--99\,076, 2019.

\bibitem{huo2018synseg}
Y.~Huo, Z.~Xu, H.~Moon, S.~Bao, A.~Assad, T.~K. Moyo \emph{et~al.},
  ``Synseg-net: Synthetic segmentation without target modality ground truth,''
  \emph{IEEE transactions on medical imaging}, vol.~38, no.~4, pp. 1016--1025,
  2018.

\bibitem{van2019learning}
G.~van Tulder and M.~de~Bruijne, ``Learning cross-modality representations from
  multi-modal images,'' \emph{IEEE transactions on medical imaging}, vol.~38,
  no.~2, pp. 638--648, 2019.

\bibitem{bilen2017universal}
H.~Bilen and A.~Vedaldi, ``Universal representations: The missing link between
  faces, text, planktons, and cat breeds,'' \emph{arXiv:1701.07275}, 2017.

\bibitem{karani2018lifelong}
N.~Karani, K.~Chaitanya, C.~Baumgartner, and E.~Konukoglu, ``A lifelong
  learning approach to brain {MR} segmentation across scanners and protocols,''
  in \emph{MICCAI}, 2018, pp. 476--484.

\bibitem{hinton2015distilling}
G.~Hinton, O.~Vinyals, and J.~Dean, ``Distilling the knowledge in a neural
  network,'' \emph{arXiv preprint arXiv:1503.02531}, 2015.

\bibitem{tzeng2015simultaneous}
E.~Tzeng, J.~Hoffman, T.~Darrell, and K.~Saenko, ``Simultaneous deep transfer
  across domains and tasks,'' in \emph{ICCV}, 2015, pp. 4068--4076.

\bibitem{hou2018lifelong}
S.~Hou, X.~Pan, C.~Change~Loy, Z.~Wang, and D.~Lin, ``Lifelong learning via
  progressive distillation and retrospection,'' in \emph{Proceedings of the
  European Conference on Computer Vision (ECCV)}, 2018, pp. 437--452.

\bibitem{papernot2016distillation}
N.~Papernot, P.~McDaniel, X.~Wu, S.~Jha, and A.~Swami, ``Distillation as a
  defense to adversarial perturbations against deep neural networks,'' in
  \emph{2016 IEEE Symposium on Security and Privacy (SP)}.\hskip 1em plus 0.5em
  minus 0.4em\relax IEEE, 2016, pp. 582--597.

\bibitem{lee2018self}
S.~H. Lee, D.~H. Kim, and B.~C. Song, ``Self-supervised knowledge distillation
  using singular value decomposition,'' in \emph{European Conference on
  Computer Vision}.\hskip 1em plus 0.5em minus 0.4em\relax Springer, 2018, pp.
  339--354.

\bibitem{wang2019segmenting}
H.~Wang, D.~Zhang, Y.~Song, S.~Liu, Y.~Wang, D.~Feng \emph{et~al.},
  ``Segmenting neuronal structure in 3d optical microscope images via knowledge
  distillation with teacher-student network,'' in \emph{ISBI}.\hskip 1em plus
  0.5em minus 0.4em\relax IEEE, 2019.

\bibitem{kats2019soft}
E.~Kats, J.~Goldberger, and H.~Greenspan, ``Soft labeling by distilling
  anatomical knowledge for improved {MS} lesion segmentation,''
  \emph{arXiv:1901.09263, ISBI}, 2019.

\bibitem{christodoulidis2016multisource}
S.~Christodoulidis, M.~Anthimopoulos, L.~Ebner, A.~Christe, and S.~Mougiakakou,
  ``Multisource transfer learning with convolutional neural networks for lung
  pattern analysis,'' \emph{IEEE journal of biomedical and health informatics},
  vol.~21, no.~1, pp. 76--84, 2016.

\bibitem{ioffe2015batch}
S.~Ioffe and C.~Szegedy, ``Batch normalization: Accelerating deep network
  training by reducing internal covariate shift,'' \emph{arXiv preprint
  arXiv:1502.03167}, 2015.

\bibitem{ulyanov2016instance}
D.~Ulyanov, A.~Vedaldi, and V.~Lempitsky, ``Instance normalization: The missing
  ingredient for fast stylization,'' \emph{arXiv preprint arXiv:1607.08022},
  2016.

\bibitem{ba2016layer}
J.~L. Ba, J.~R. Kiros, and G.~E. Hinton, ``Layer normalization,'' \emph{arXiv
  preprint arXiv:1607.06450}, 2016.

\bibitem{wu2018group}
Y.~Wu and K.~He, ``Group normalization,'' in \emph{Proceedings of the European
  Conference on Computer Vision (ECCV)}, 2018, pp. 3--19.

\bibitem{cciccek20163d}
{\"O}.~{\c{C}}i{\c{c}}ek, A.~Abdulkadir, S.~S. Lienkamp, T.~Brox, and
  O.~Ronneberger, ``3d u-net: learning dense volumetric segmentation from
  sparse annotation,'' in \emph{International conference on medical image
  computing and computer-assisted intervention}.\hskip 1em plus 0.5em minus
  0.4em\relax Springer, 2016, pp. 424--432.

\bibitem{milletari2016v}
F.~Milletari, N.~Navab, and S.-A. Ahmadi, ``V-net: Fully convolutional neural
  networks for volumetric medical image segmentation,'' in \emph{2016 Fourth
  International Conference on 3D Vision (3DV)}.\hskip 1em plus 0.5em minus
  0.4em\relax IEEE, 2016, pp. 565--571.

\bibitem{zhuang2019evaluation}
X.~Zhuang, L.~Li, C.~Payer, D.~Stern, M.~Urschler, M.~P. Heinrich
  \emph{et~al.}, ``Evaluation of algorithms for multi-modality whole heart
  segmentation: An open-access grand challenge,'' \emph{preprint
  arXiv:1902.07880}, 2019.

\bibitem{landman2015multi}
B.~Landman, Z.~Xu, J.~Igelsias, M.~Styner, T.~Langerak, and A.~Klein, ``2015
  miccai multi-atlas labeling beyond the cranial vault – workshop and
  challenge,'' 2015, https://www.synapse.org/\#!Synapse:syn3193805/wiki/89480.

\bibitem{payer2017multi}
C.~Payer, D.~{\v{S}}tern, H.~Bischof, and M.~Urschler, ``Multi-label whole
  heart segmentation using cnns and anatomical label configurations,'' in
  \emph{International Workshop on Statistical Atlases and Computational Models
  of the Heart}.\hskip 1em plus 0.5em minus 0.4em\relax Springer, 2017, pp.
  190--198.

\bibitem{moeskops2016deep}
P.~Moeskops, J.~M. Wolterink, B.~H. van~der Velden, K.~G. Gilhuijs, T.~Leiner
  \emph{et~al.}, ``Deep learning for multi-task medical image segmentation in
  multiple modalities,'' in \emph{MICCAI}, 2016, pp. 478--486.

\end{thebibliography}

\end{document}